\documentclass[10pt,twocolumn,letterpaper]{article}

\usepackage{cvpr}              %

\newcommand{\myparagraph}[1]{\vspace{0.1em}\noindent\textbf{#1}}

\usepackage{xcolor}

\usepackage{url}
\usepackage{adjustbox}
\usepackage{lineno}
\usepackage{graphicx}
\usepackage{booktabs}
\usepackage{wrapfig}
\usepackage{float}
\usepackage{colortbl}

\definecolor{cvprblue}{rgb}{0.21,0.49,0.74}
\usepackage[pagebackref,breaklinks,colorlinks,allcolors=cvprblue]{hyperref}
\usepackage[dvipsnames]{xcolor}

\def\paperID{*} %
\def\confName{CVPR}
\def\confYear{2026}

\title{CGHair: Compact Gaussian Hair Reconstruction with Card Clustering}

\author {
    Haimin Luo$^{1,2}$ \quad Srinjay Sarkar$^{1}$ \quad Albert Mosella-Montoro$^{1}$ \\ Francisco Vicente Carrasco$^{1}$ \quad Fernando De la Torre$^{1}$\\ [6pt]
    \textsuperscript{1} Carnegie Mellon University \quad \textsuperscript{2} ShanghaiTech University\\ [6pt]
    \url{https://humansensinglab.github.io/CGHair/}
}

\begin{document}
\twocolumn[{%
\renewcommand\twocolumn[1][]{#1}%
\maketitle
\begin{center}
    \centering
    \captionsetup{type=figure}
    \vspace{-6mm}
    \includegraphics[width=1.0\textwidth]{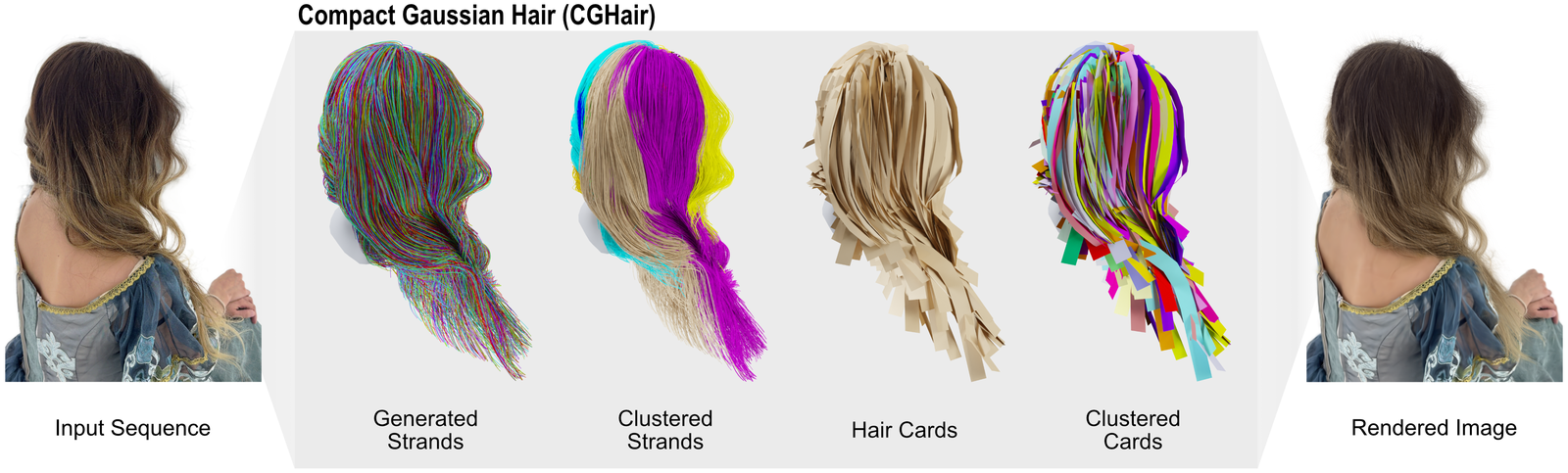}
    \vspace{-6mm}
    \caption{
Our method introduces a novel hierarchical clustering and encoding framework for modeling human hair. From an input sequence, we extract strand-level hair geometry and cluster it into representative hair cards, each sharing a Gaussian texture codebook to encode appearance. This representation, termed CGHair (Compact Gaussian Hair), achieves remarkably high compression of hair appearance while maintaining high-quality free-view rendering.    } 
    \label{fig:fig_teaser}
\end{center}%
}]

\begin{abstract}
We present a compact pipeline for high-fidelity hair reconstruction from multi-view images. While recent 3D Gaussian Splatting (3DGS) methods achieve realistic results, they often require millions of primitives, leading to high storage and rendering costs. Observing that hair exhibits structural and visual similarities across a hairstyle, we cluster strands into representative hair cards and group these into shared texture codebooks. Our approach integrates this structure with 3DGS rendering, significantly reducing reconstruction time and storage while maintaining comparable visual quality. In addition, we propose a generative prior accelerated method to reconstruct the initial strand geometry from a set of images. 
Our experiments demonstrate a 4-fold reduction in strand reconstruction time and achieve comparable rendering performance with over 200× lower memory footprint.

\end{abstract}

\let\origaddcontentsline\addcontentsline
\renewcommand{\addcontentsline}[3]{} %

\section{Introduction}
\label{sec:intro}

Human hair is an inherently distinctive and visually striking aspect of personal appearance, making its accurate reconstruction crucial for the realistic depiction of human avatars. Early research in this field primarily addressed the challenge of reconstructing human hair geometry at the strand level, laying the groundwork for subsequent advancements~\citep{kuang2022deepmvshair, nam2019strand, rosu2022neural, wang2004hair, Hu2015SingleHairData, autohair, Paris2008HairPhotobooth}.
Recent advances have enabled high-fidelity reconstruction of dense hair geometry and realistic appearance directly from multi-view images by combining 3D Gaussian Splatting (3DGS) with strand-based modeling~\citep{sklyarova2023neural, luo2024gaussianhair, zhou2024groomcap, zheng2025groomlight}. State-of-the-art methods (e.g., GaussianHair~\cite{luo2024gaussianhair}) use chains of cylindrical Gaussians to better capture strand-like structures and jointly optimize appearance and lighting parameters. However, modeling highly dense hair still requires millions of primitives, leading to substantial redundancy, storage, and rendering challenges that limit practical adoption.
Hair geometry and appearance are often consistent across a hairstyle—a feature harnessed by hair card representations to reduce the computational cost of animating and storing hair geometry and appearance. Building on this principle, we introduce a compact Gaussian-based hair reconstruction pipeline from multi-view images, guided by hierarchical clustering techniques inspired by hair cards. 

\begin{figure*}[t!]
	\includegraphics[width=\linewidth]{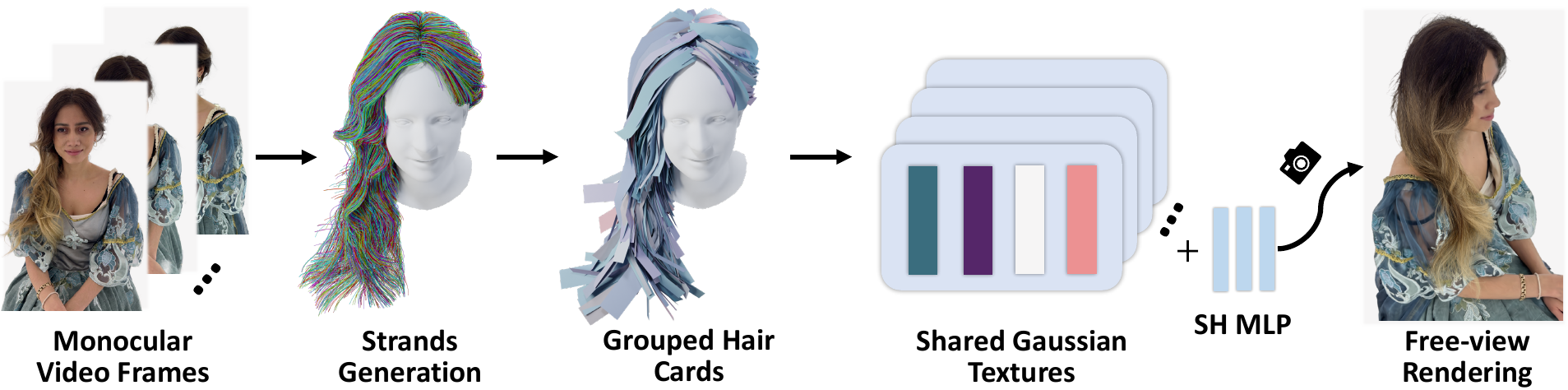}
        \vspace{-20pt} 
	\caption{\textbf{Full pipeline.} Given monocular video frames, we first reconstruct hair strands with our efficient strand generator. The strands are grouped by hair cards, which are further clustered into card groups. Each card group is assigned to a shared Gaussian Texture for compact appearance modeling, a specialized global MLP is used to compute Spherical Harmonics (SH) from the Gaussian Textures.}
        \vspace{-15pt} 
	\label{fig:pipeline}
\end{figure*}

Our method rapidly reconstructs hair strands from images using a generative, prior-driven approach, then groups them by geometry and creates representative hair cards for each cluster. These cards are further organized by appearance into clusters with shared Gaussian texture codebooks, enabling unified appearance attributes, as illustrated in ~\cref{fig:fig_teaser}. The resulting structure integrates seamlessly into a 3DGS pipeline, employing specialized optimization for efficient multi-view reconstruction. Experiments demonstrate significantly reduced reconstruction time and competitive visual quality with 200-fold less memory footprint.
In summary, our main contributions are:

\begin{itemize}
    \item We develop a generative prior accelerated method for efficiently reconstructing strand-level hair geometry from multi-view images.
    
    \item We design a compact hair modeling pipeline that leverages hair card-guided hierarchical clustering to significantly reduce redundancy at the strand level.
    
    \item We propose a shared Gaussian appearance codebook for hair cards, enabling scalable and consistent appearance modeling across structurally similar hair regions.
\end{itemize}

\section{Related Work}
\label{sec:related_work}

\myparagraph{Geometric Hair Modeling.}
Various parametric representations have been explored in early works for hair shape modeling, such as 2D parametric surfaces~\citep{koh2000real, liang2003enhanced, noble2004modelling}, cylinders~\citep{chen1999system, yang2000cluster, xu2001v, patrick2004modelling, kim2002interactive, wang2004hair, choe2005statistical}, or hair meshes~\citep{yuksel2009hair}. In contrast, hair cards are widely favored in the existing asset production pipeline. It is due to their compact nature, which comes from their shared geometry and appearance texture.
On the other hand, numerous earlier works have focused on reconstructing strand-level hair geometry from multi-view images, like volume-based methods ~\citep{grabli2002image, paris2004capture}, or orientation-based methods with high-end acquisition systems ~\citep{wei2005modeling, luo2012multi, luo2013wide, paris2008hair, lightinghair, bao2016realistic}. More recent data-driven ~\citep{hu2014robust, yu2014hybrid, zhang2017data, hu2015single} and learning-based advances ~\citep{wu2022neuralhdhair, kuang2022deepmvshair, zhou2018hairnet, saito20183d, yang2019dynamic, rosu2022neural, sklyarova2023neural, autohair, 4viewhair, zheng2023hairstep, alexandru2025difflocks, sklyarova2025im2haircut} have been proposed to infer high-quality strand geometry from single or multiple views. The development of such approaches has been enabled by a high-fidelity dataset of hair strands, as well as the high-quality geometric priors extracted from it.

\myparagraph{Neural Hair Representations.}
Besides explicit geometry, various works have introduced neural representation in hair modeling, such as neural orientation field ~\citep{wu2022neuralhdhair, kuang2022deepmvshair, sklyarova2023neural}, neural point clouds ~\citep{Wang2020NOPC}, and neural radiance field~\citep{mildenhall2021nerf} and its variants ~\citep{convnerf, Artemis, wang2022hvh, wang2023neuwigs, wu2024monohair, zhou2024groomcap}. These methods either adopt slow but compact implicit representations or explicitly discretized variants (e.g., voxels) to pursue faster rendering at the cost of significant parameter redundancy.
The recent work 3D Gaussian Splatting (3DGS)~\citep{kerbl3Dgaussians}, brought revolutionary development in both scene reconstruction quality and real-time speed. It represents scenes as a mixture of volumetric 3D Gaussians, with spherical harmonics parameters for appearance modeling. 
GaussianHair is the first 3DGS-based hair representation by reformulating a hair strand as a sequence of cylindrical Gaussians. It supports real-time rendering, animation, and re-lighting. However, it also leads to millions of Gaussians and substantial parameter redundancy. Subsequent works ~\citep{zhou2024groomcap, zakharov2024human, zheng2025groomlight, Pan_2025_BMVC} adopting similar structures share the same limitation.
In this work, we address this issue by introducing a compact 3DGS hair representation with a hierarchical hair card structure to reduce parameter redundancy.

\myparagraph{Compact 3D Neural/Gaussian Model.}
Compression techniques, e.g., vector quantization (VQ) ~\citep{gray1984vector, gersho2012vector, equitz2002new, gong2014compressing}, have been extensively explored in image and video codecs ~\citep{cosman1993using, lee1995motion}. The progressive application, like VQVAE~\citep{van2017neural}, DVAE~\citep{ramesh2021zero}, has further contributed to the emergence of large generative models such as Stable Diffusion~\citep {rombach2022high} and Sora~\citep {liu2024sora}. 
Recent works have investigated compressing general 3D neural representations, e.g., neural radiance fields~\citep{mildenhall2021nerf}. Their primary goal is to reduce the storage cost imposed by explicit acceleration structures like 3D grids~\citep{takikawa2022variable, rho2023masked, NGP, tang2022compressible}, via various techniques such as factorization~\citep{chen2022tensorf}, hash encoding~\citep{NGP}, and vector quantization~\citep{li2023compressing}. 
Recent works on 3DGS~\citep{kerbl3Dgaussians} focus on either reducing Gaussian point cloud complexity~\citep{girish2024eagles, fan2024lightgaussian, niedermayr2024compressed, papantonakis2024reducing, lee2024compact} or compressing Gaussian attributes to lower storage overhead~\citep{lee2024compact, morgenstern2024compact, liu2024compgs, chen2024hac, lee2025compression, navaneet2023compact3d}.
While prior studies have made notable progress, they fail to exploit redundancies within Gaussian attributes—a critical omission for hair, where geometry and appearance exhibit strong structural regularity. We propose a card-based hierarchical representation that organizes strands into hair cards and employs shared appearance codebooks. By explicitly modeling appearance redundancy across similar strand clusters, our method achieves high compression while preserving compatibility with real‑time Gaussian-based rendering.

\section{Method}
\label{sec:method}

This section presents the proposed techniques to reconstruct our compact CGHair from a monocular video, as shown in ~\cref{fig:pipeline}. First, high-fidelity hair strands are generated using our generative prior–accelerated method (Sec.\ref{sec:strand_generator}); Subsequently, the strands are grouped into representative hair cards with extracted surface textures (Sec.\ref{sec:hair_card}), which are further clustered into a shared Gaussian texture for compact Gaussian Hair representation (\cref{sec:compact_gh}). This representation is then integrated into the 3DGS pipeline with a tailored appearance optimization scheme (\cref{sec:appearance_modeling}).

\subsection{Hair Strand Generation} \label{sec:strand_generator}

Building on prior Gaussian-based approaches, such as GaussianHaircut and the generative PCA prior introduced in PERM~\citep{he2024perm}, we present an efficient framework that achieves a four-fold acceleration.

\myparagraph{Preprocessing.}
Given a sequence of multi-view images, we first estimate the corresponding camera poses and compute hair segmentation masks for each frame with the approach of~\citet{yao2024matte}. Next, we derive hair orientation maps by applying a bank of Gabor filters, which provide robust local directional cues for subsequent strand reconstruction.

\myparagraph{Gaussian Fitting.}
We reconstruct two distinct Gaussian sets: one representing the head and another corresponding to the hair region, following ~\citet{zakharov2024human}. The head Gaussians remain fixed in the subsequent stages, while the hair Gaussians are modeled as cylindrical primitives to capture strand-like geometry. The rotation of these cylindrical Gaussians is guided by 2D orientation maps, providing refined supervision for surface alignment during hair reconstruction. Finally, a FLAME head model~\citep{FLAME:SiggraphAsia2017} is fitted to the head Gaussians for use in the next processing stage.

\myparagraph{Strand Generation.}
Given the scalp mesh from the fitted FLAME head model, along with multi-view RGB and orientation images, we reconstruct a dense set of hair strands.

The hair strand geometry is represented as a latent hair texture parameterized on the scalp mesh. Directly optimizing these latent textures from scratch is inefficient, as it requires strong structural priors or computationally heavy diffusion-based regularization to maintain plausible strand geometry.
\begin{figure}
  \includegraphics[width=1.0\linewidth]{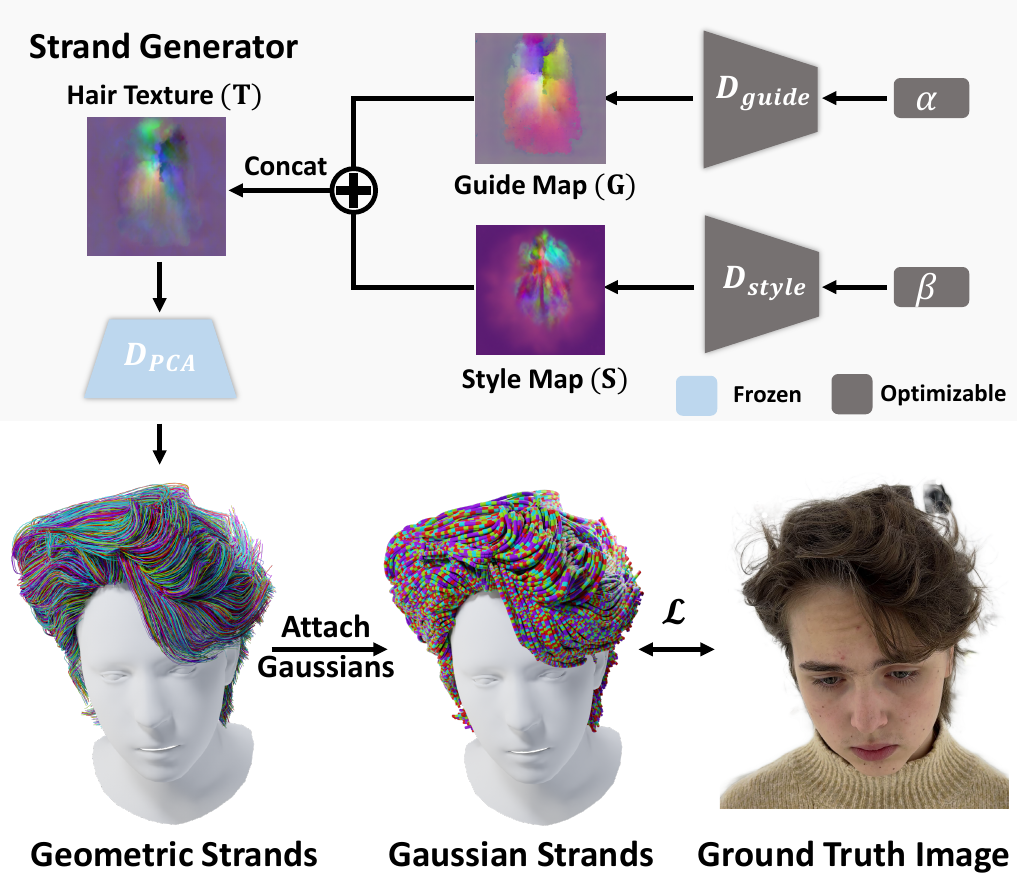}
  \vspace{-20pt} 
  \caption{ \textbf{Strand Generation Pipeline.} We use the parametric hair geometric model (PERM) to synthesize a hair texture and decode it into strands. We then attach cylindrical Gaussian primitives to the strands and optimize them from images using the 3DGS.}
  \label{fig:strand_generator}
  \vspace{-15pt}
\end{figure}
To address this challenge, we adopt the parametric geometric hair model PERM~\citep{he2024perm}, which encodes robust priors on hair structure and local strand variability. Thus enabling efficient, stable, and image-driven reconstruction without slow latent-space optimization or costly loss functions. Following PERM, we represent hair as a 2D latent UV texture map defined on the scalp mesh, disentangled into a guide map $\mathbf{G}$ capturing global hair structure and a style map $\mathbf{S}$ capturing local strand details. As illustrated in ~\cref{fig:strand_generator}, we exploit the PERM decoders, pretrained on a complex synthetic hairstyle dataset~\cite{he2024perm}: a StyleGAN2-based guide decoder $\mathcal{D}_{guide}$ and a VAE-based style decoder $\mathcal{D}_{style}$. These decoders generate the maps from $\alpha$, guide code, and $\beta$, style code, as $\mathbf{G} = \mathcal{D}_{guide}(\alpha)$ and $\mathbf{S} = \mathcal{D}_{style}(\beta)$. The resulting maps are concatenated into a geometric hair texture $\mathbf{T}$, whose texels correspond to PCA coefficients later decoded into full 3D strands by a pretrained lightweight PCA decoder $\mathcal{D}_{\mathrm{PCA}}$~\citep{he2024perm}. Next, we attach cylindrical Gaussians along the reconstructed strands and integrate them into the 3DGS differentiable rendering framework. Each line segment of a strand is represented by a Gaussian whose length matches the segment length and orientation aligns with the local tangent direction. Each Gaussian is further associated with trainable spherical harmonic coefficients for appearance modeling, allowing the photometric supervision to refine the geometric structure. All Gaussians are treated as opaque to prevent blending artifacts and ambiguities from opacity. Finally, we jointly optimize the hairstyle codes $\alpha$ and $\beta$ and fine-tune the pretrained decoders $\mathcal{D}_{guide}$ and $\mathcal{D}_{style}$ using photometric and orientation losses~\citep{zakharov2024human} between rendered and input views. Leveraging the strong geometric prior of PERM and the compact PCA-based strand reconstruction, our framework achieves high-fidelity strand generation with substantially reduced computation time compared to previous methods.

\subsection{Hair Card Generation.} \label{sec:hair_card}

Hair cards efficiently approximate small hair bundles with polygonal strip meshes rather than individual strands, encoding bundle shape, volume, and orientation via shareable textures. We convert our reconstructed strand into hair cards by clustering strands, constructing a strip-based card for each cluster, and generating per-card geometry textures from the strands for our compact hair representation.

\myparagraph{Strand Clustering.}
Given reconstructed strands where a strand is a 3D polyline composed of a sequence of points, i.e., $\mathcal{S} = \{p_i\}, p_i =(x_i,y_i,z_i)$, we concatenate each strand’s points into a vector and group them into $N_c$ clusters using k-means algorithm with $l_2$ as distance metric. The centroid of each cluster is used to guide a strand that captures the cluster’s overall flow.

\begin{figure}
  \centering
  \includegraphics[width=1.0\linewidth]{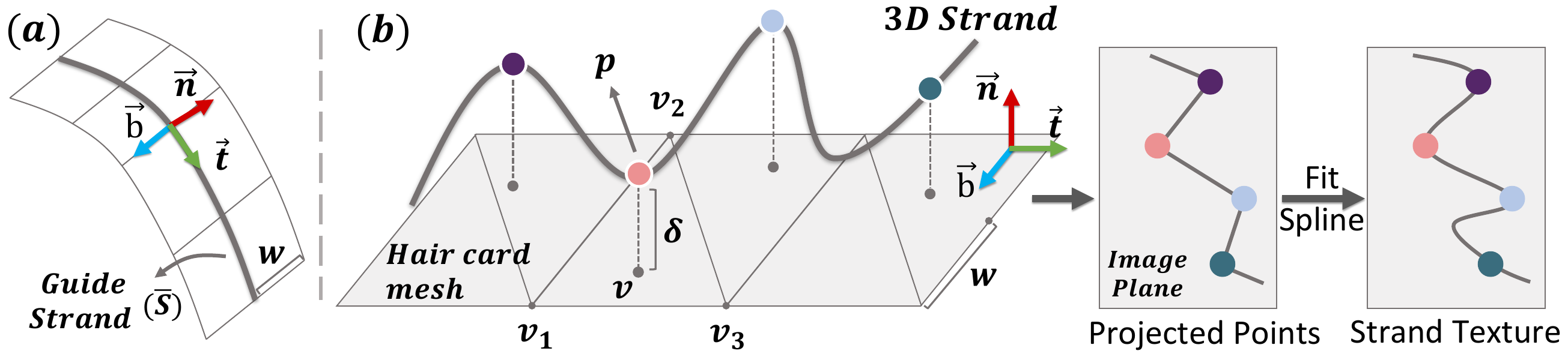}
  \vspace{-20pt} 
  \caption{(a) Hair card geometry construction; (b) Strand texture generation.}
  \label{fig:card_based_hair}
  \vspace{-10pt} 
\end{figure}

\myparagraph{Hair Card Geometry Construction.}
We build hair cards from the hair strand clusters and corresponding guide strands. To ensure clarity, we present the process for a single cluster $\mathcal{C} = \{\mathcal{S}_l\}$. The guide strand $\bar{\mathcal{S}}$ is fitted by a B-spline curve and subsequently smoothed and downsampled as $\bar{\mathcal{S}} = \{ \mathbf{\bar p}_k\}$ to reduce the card geometric complexity. 
As shown in ~\cref{fig:card_based_hair} (a), $\bar{\mathcal{S}}$ serves as the central axis of the card, and the geometry is constructed by extending in both directions along the bitangent vector $\mathbf{b}_k$, which lies in the plane orthogonal to the strand direction $\mathbf{t}_k$ and normal vectors $\mathbf{n}_k$.
Based on this formulation, constructing the card primarily involves two steps: 1) Determine the normal vector for each segment of the guide strand; 2) Set the card width along the bitangent direction according to the maximum distance between the guide strand and other strands in the cluster, ensuring all strands fit within the card.

Note that the normals strictly lie in the plane orthogonal to $\mathbf{t}_k$; Thus, we define the estimation of the normals as a constrained optimization process to minimize the sum of absolute distances between strand points and the card plane:
\begin{equation}
\begin{aligned}
    \{\mathbf{n}_k^\ast\} \;=\; \arg\min_{\{\mathbf{n}_k\}} \; 
    & \sum_k \sum_{\mathbf{p}_i \in \mathcal{N}_k} 
      \left\| (\mathbf{p}_i - \bar{\mathbf{p}}_k) \cdot \mathbf{n}_k \right \| \quad   \\
    \text{s.t. } %
    & \mathbf{n}_k \cdot \mathbf{t}_k = 0, \quad \|\mathbf{n}_k\| = 1, \quad \forall k.
\end{aligned}
\end{equation}

where $\mathcal{N}_k$ contains the points from all strand segments in the cluster that overlap with the $k$-th center strand segment. The normals are optimized differentiably via gradient descent, after which adjacent normals with opposite directions are flipped, and Gaussian smoothing is applied to enforce orientation consistency.

With the optimized normals, the bitangent can then be defined by $\mathbf{b}_k = \mathbf{n}_k \times \mathbf{t}_k$, and the card width $w_k$ is defined as:
\begin{equation}
    w_k = \max_k\max_{\mathbf{p}_i\in \mathcal{N}_k} \left \| (\mathbf{p}_i - \mathbf{\bar p}_k) \cdot \mathbf{b}_k \right \|.
\end{equation}
the card vertices are then calculated as $\mathbf{\bar p}_k \pm w_k\mathbf{b}_k$,  and sequentially connected to recover the hair card mesh.

\begin{figure*}[t!]%
\vspace{-20pt} 
\centering
    \includegraphics[width=1.0
    \linewidth]{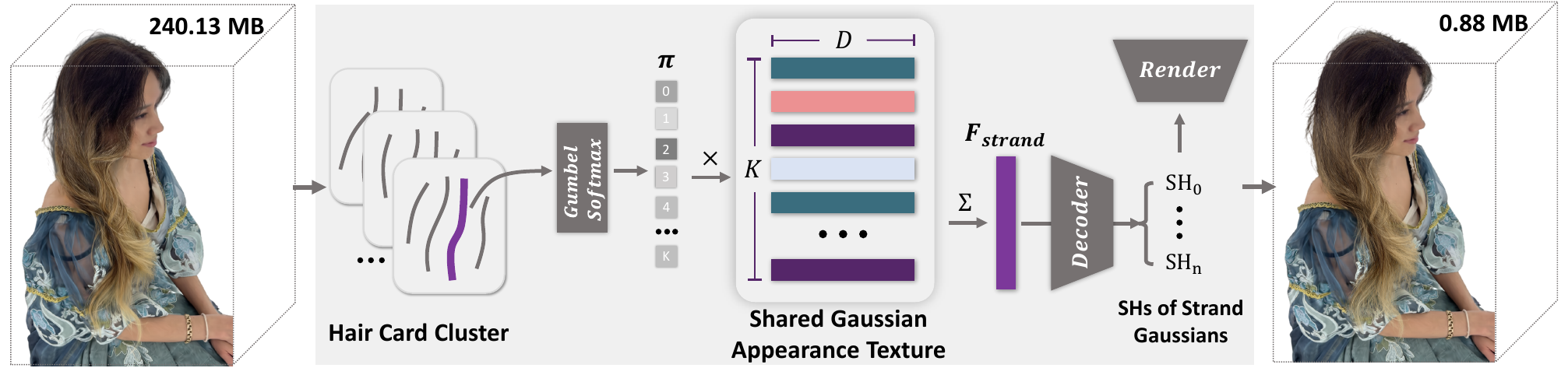}
    \vspace{-20pt} 
  \caption{We model each hair card cluster with a shared Gaussian appearance texture and a lightweight globally-shared decoder, achieving over 270-fold compression.}
  \label{fig:appearnce_modeling}
  \vspace{-15pt} 
\end{figure*}

\myparagraph{Strand Texture Generation.}
Given the hair card geometry, we generate 2D strand textures $\{\mathcal{T}_i\}$ by projecting the 3D strand polylines onto the card surface to establish a UV mapping. However, directly projecting reconstructed strands often introduces ambiguities such as overlaps and twists due to their complex spatial configuration. To mitigate these issues, we employ an optimization-based inverse mapping procedure that robustly estimates the UV correspondence between strand geometry and the card surface.
For a strand in a cluster $\mathcal{C}$, we initialize a set of optimizable 2D strand coordinates $\{\mathbf{u}_i\}$ in UV space. For each point $\mathbf{u}_i$, and the triangle $\mathcal{T}_i = \{\mathbf{u}_{i,1}, \mathbf{u}_{i,2}, \mathbf{u}_{i,3}\}$ it lies in, the barycentric coordinates $(\lambda_1, \lambda_2, \lambda_3)$ are determined from
$\mathbf{u}_i = \sum_{m=1}^{3} \lambda_m \mathbf{u}_{i,m}$, where$\sum_{m=1}^{3} \lambda_m = 1$.
As shown in ~\cref{fig:card_based_hair}(b), these coordinates are then used to map $\mathbf{u}_i$ to 3D points with 3D triangle vertex positions $(\mathbf{v}_{i,1}, \mathbf{v}_{i,2}, \mathbf{v}_{i,3})$ and an additional learnable displacement $\delta_i$ along the interpolated surface normal $\mathbf{\hat{n}}_i$ :
\begin{equation} \label{eq:strand_representation}
    \mathbf{\hat{p}}_i = \mathbf{v}_i + \delta_i\mathbf{\hat{n}}_i, \quad \mathbf{v}_i = \sum_{m=1}^{3}{\lambda_m \mathbf{v}_{i,m}}
\end{equation} 
A $l_2$ loss is then applied to jointly optimize $\{\mathbf{u}_i\}$ and $\{\delta_i\}$:
\begin{equation}
    \mathcal{L}_{strand} = \sum_i \| \mathbf{\hat{p}}_i - \mathbf{p}_i \|^2
\end{equation}
Following the optimization, we obtain discrete 2D strand point sets in UV space. To generate high-quality, anti-aliased textures, each strand is modeled as a smooth 2D B-spline curve fitted to its UV coordinates. These curves are densely upsampled to produce sequential point samples, then rasterized onto the image plane (~\cref{fig:card_based_hair}(b)). The resulting strand texture map encodes the geometric structure of the card's associated strands, providing a compact and smooth representation for downstream processing.

\subsection{Compact Gaussian Hair Representation Based on Card Clustering.} \label{sec:compact_gh}

Using the generated hair cards and their associated strand textures, we construct a compact Gaussian-based hairstyle representation. The geometry of the Gaussian hair is defined directly by the hair cards. For appearance modeling, we cluster hair cards based on their strand textures and assign a shared Gaussian appearance texture to each cluster. This approach encodes all strands associated with the clustered cards, enabling an efficient compression of appearance information while preserving strand characteristics.

\myparagraph{Card-Based Gaussian Strand Geometry.}
As derived from ~\cref{eq:strand_representation}, strand points are parameterized using barycentric coordinates \(\mathbf{b}_i\) relative to their corresponding hair card mesh faces, alongside normal-direction displacements \(\delta_i\). Cylindrical Gaussians are positioned along each line segment \(\{\mathbf{p}_i, \mathbf{p}_{i+1}\}\) to form the Gaussian strand representation. Specifically, each Gaussian is centered at \(\mathbf{\mu}_i = (\mathbf{p}_i + \mathbf{p}_{i+1}) / 2\), with a scale defined by \(\{d, d, \frac{s}{2}\}\), where \(s\) is the segment length between adjacent strand points and \(d\) is a small constant scale. The rotation matrix aligns the principal axis with the segment direction \(\mathbf{p}_{i+1} - \mathbf{p}_i\). Incorporating per-Gaussian opacity parameters, the entire Gaussian strand geometry is compactly represented by the sets \(\{\mathbf{b}_i\}^l\), \(\{\delta_i\}^l\), and \(\{\sigma_i\}^{l-1}\), where \(l\) denotes the number of strand points.

\myparagraph{Shared Gaussian Appearance Texture.}
For the appearance of hair Gaussians, we introduce a compact Shared Gaussian Appearance Texture that clusters the appearances of associated strands into common textures. This approach leverages the structural and visual redundancies inherent in hair, enabling efficient compression of spherical harmonic coefficients—the main source of appearance redundancy in Gaussian hair representations.
To identify hair cards with similar structures, we cluster them based on their strand texture features. Specifically, features are extracted from the strand textures $\{\mathcal{T}_i\}$ described in ~\cref{sec:hair_card},
and k-means clustering is applied to group the textures into \(N_T\) clusters according to their feature similarity. For each hair card cluster, we randomly initialize an appearance codebook with \(K\) features,
as illustrated in ~\cref{fig:appearnce_modeling}, which are updated during subsequent training and serve as shared textures for all strands represented by the cards within that cluster. Each codebook entry encodes the complete view-dependent appearance of a strand, including spherical harmonic coefficients across its Gaussians, enabling a compact yet expressive representation. Since \(K\) is significantly smaller than the number of associated strands, this method achieves a substantial compression ratio.

\begin{figure*}[t!]
        \vspace{-20pt} 
	\includegraphics[width=0.95\linewidth]{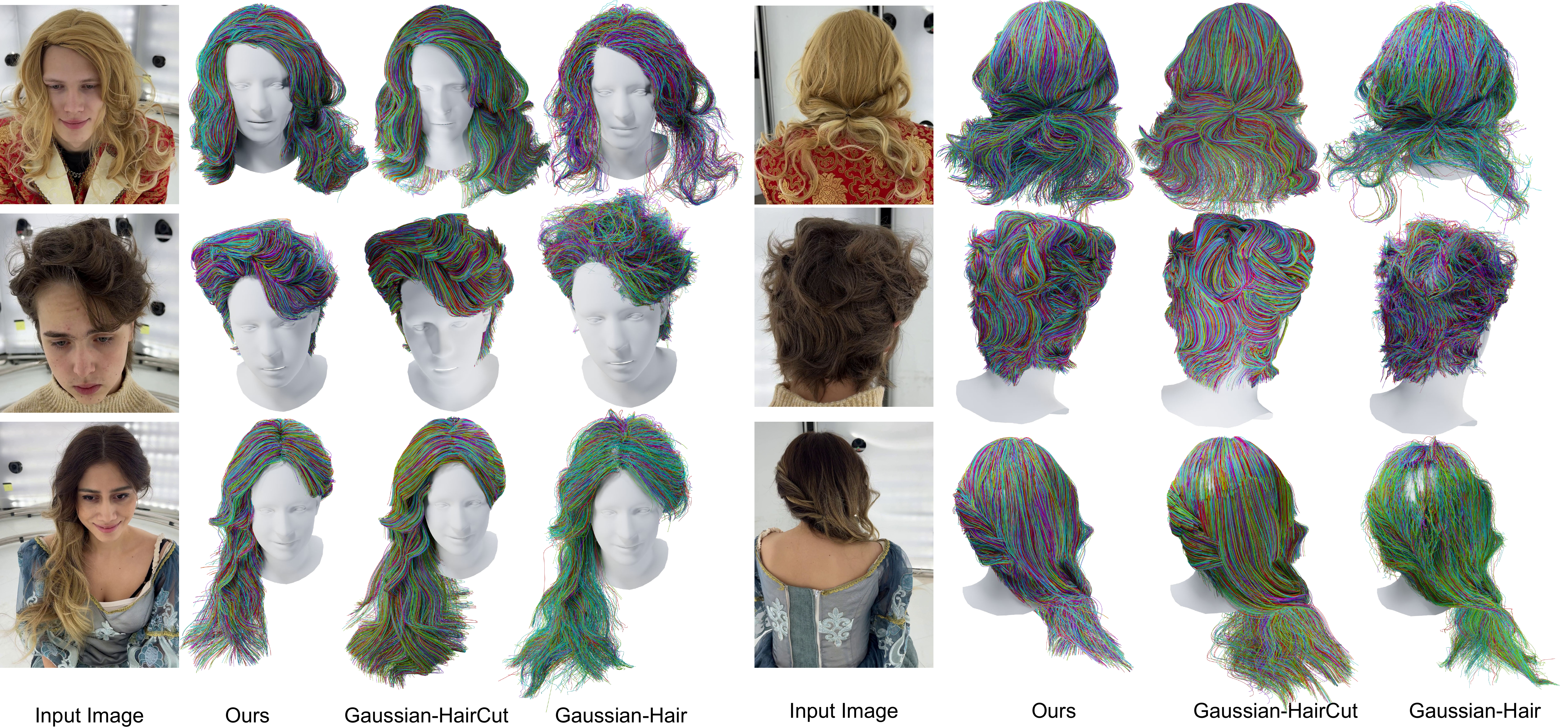   }
        \vspace{-8pt} 
	\caption{Qualitative comparison of reconstructed hair strands with GaussianHaircut and GaussianHair. Our method achieves higher geometric fidelity, over 4× speedup (vs. GaussianHaircut~\cite{zakharov2024human}), and a fully automatic pipeline (unlike GaussianHair~\cite{luo2024gaussianhair}).}
	\label{fig:comparison}
        \vspace{-10pt} 
\end{figure*}

\subsection{Hair Appearance Modeling from Images. }\label{sec:appearance_modeling}
Based on our constructed compact Gaussian Hair representation, we present a scheme that optimizes this representation from multi-view images and reconstructs the hair appearance. 
Specifically, we formulate the hair appearance modeling at the strand level, in contrast to prior approaches that operate at the level of individual Gaussian primitives, as shown in ~\cref{fig:appearnce_modeling}.
Moreover, for each strand in the hairstyle, we introduce an additional learnable $K$-dimensional logit vector, which serves as a soft index into the corresponding appearance codebook.  
Specifically, the strand is first routed to its assigned codebook according to its card and corresponding card cluster. 
The logit is then passed through a Gumbel Softmax~\citep{jang2016categorical} operation to produce a soft label, which is then used to compute a weighted combination of all $K$ codebook entries:
\begin{equation}
    \mathbf{F}_\text{strand} = \sum_{k=1}^{K} \pi_k \cdot \mathbf{F}_k,
\end{equation}
where $\pi_k $ denotes the soft assignment weight for the $k$-th entry, $ \mathbf{F}_k $ is the corresponding feature from the codebook.

The resulting feature $\mathbf{F}_\text{strand}$ can be directly interpreted as all the SH coefficients of the Gaussian along the strand, and applied in the 3DGS rendering pipeline.
However, in the common case of a hair strand containing, e.g., 100 Gaussians, such a straightforward method leads to substantial redundancy, as each codebook entry must capture a high-dimensional feature (e.g., over 4k dimensions). This significantly increases the codebook storage footprint.
Besides, considering the nature of a hairstyle generally contains tens of thousands of strands, this design further amplifies the memory and computational cost during the optimization.  

To mitigate this problem, we draw inspiration from the paradigm of modern quantization-based methods ~\citep{van2017neural, ramesh2021zero} and perform the sharing of Gaussian appearance texture in a more compact feature space. 
Therefore, the appearance feature dimensionality $D$ is set to be significantly smaller (e.g., $ D = 64 $) to ensure a compact representation.  
The resulting strand-level feature $ \mathbf{F}_\text{strand} $ is subsequently decoded by a globally shared MLP $\phi_\text{dec}$ into the independent SH parameters $\mathbf{SH}_i$ for each Gaussian along the strand $\mathcal{S}$:
\begin{equation}
    [\mathbf{SH}_1, \mathbf{SH}_2, \dots, \mathbf{SH}_{|\mathcal{S}|}] = \phi_\text{dec}(\mathbf{F}_\text{strand})
\end{equation}

\myparagraph{Optimization.}
Our model can be optimized in an end-to-end fashion using multi-view supervision to reconstruct compact and faithful Gaussian hair from input images.
We adopt the original photometric loss $\mathcal{L}_{p} = \mathcal{L} + \mathcal{L}_{D-SSIM}$ in 3DGS for appearance reconstruction.
We also adopt the alpha loss $\mathcal{L}_a$ between the rendered alpha map and alpha supervision, and another opacity smoothness regularization term $\mathcal{L}_o$ from ~\citet{luo2024gaussianhair}.
The total loss function is:
\begin{equation}
    \mathcal{L} = \mathcal{L}_p + \mathcal{L}_a + \mathcal{L}_o
\end{equation}

\section{Experiments}
\label{sec:experiment}
\subsection{ Implementation Details.} \label{sec:implementation}
\myparagraph{Dataset.} 
We use three challenging real-world hairstyles (curly, short, long)~ for comparison from the GaussianHair~\cite{luo2024gaussianhair} dataset, training on about 100 randomly and uniformly sampled frames and evaluating novel-view rendering on the rest. 
For ablations, we use several synthetic hairstyles from HairSalon dataset~\cite{hu2015single} and render monocular videos, providing accurate strand geometry and cameras for quantitative evaluation. Refer to the supplementary materials for more evaluations on MonoHair dataset~\cite{wu2024monohair}, simulations, and discussions.

\begin{figure*}[t!]
        \vspace{-20pt} 
	\includegraphics[width=\linewidth]{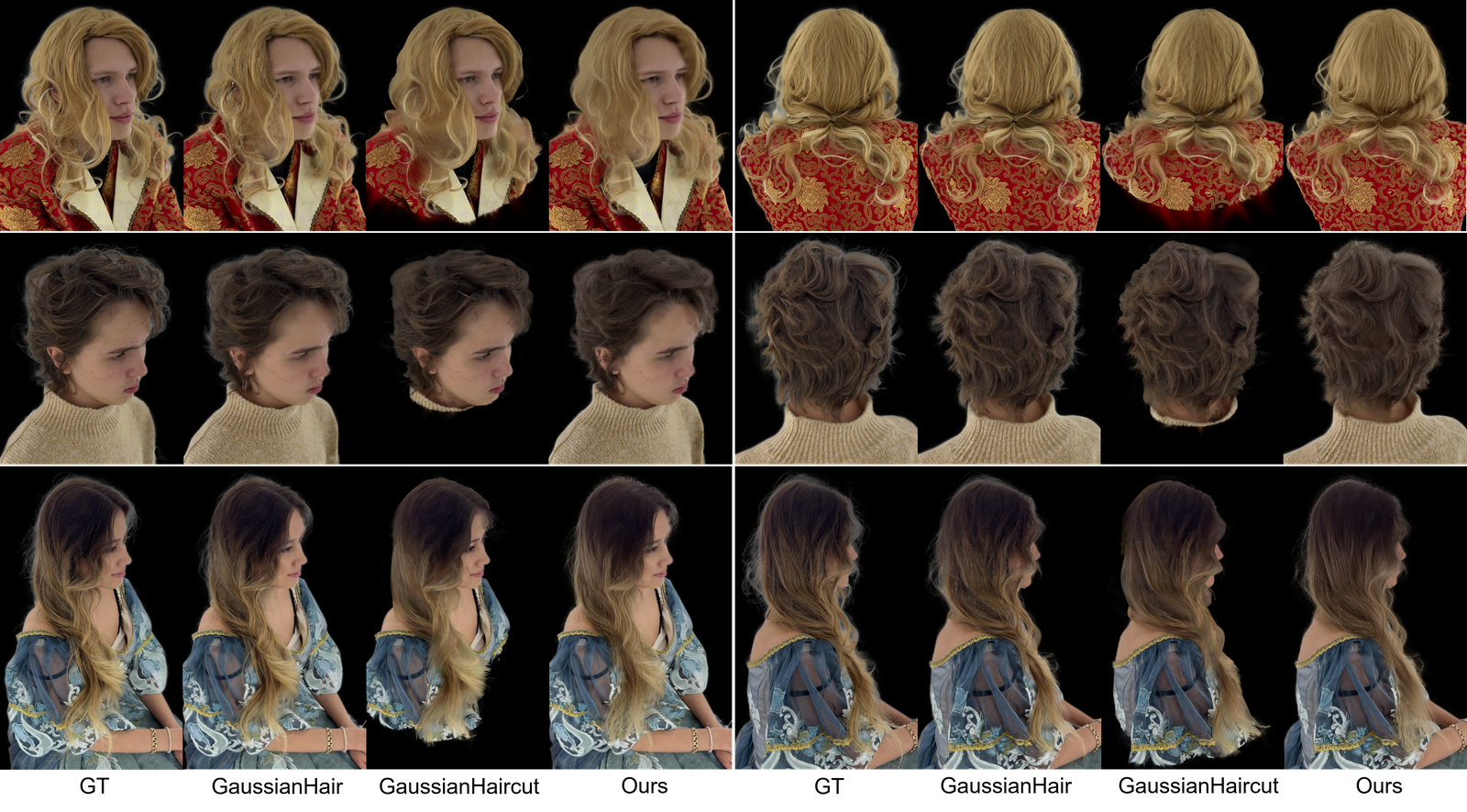}
        \vspace{-20pt} 
	\caption{Novel-view rendering comparisons on three challenging real-world hairstyles against GaussianHair~\cite{luo2024gaussianhair} and GaussianHaircut~\cite{zakharov2024human}. CGHair achieves comparable quality with 200x fewer parameters. Refer to the supplementary material for quantitative results.}
	\label{fig:appearance_comp}
\vspace{-10pt}
\end{figure*}

\myparagraph{Implementation.}  
We train the strand generator in two stages: Gaussian fitting and strand generation. In the Gaussian fitting stage, we follow the base 3DGS hyperparameters and optimize for 7k iterations with Adam~\citep{kingma2014adam}. For strand generation, we initialize with the pre-trained prior of ~\citet{he2024perm}; the guide and residual texture generators produce strands from scalp roots to initialize the hair Gaussians, and an appearance decoder predicts per-Gaussian spherical harmonics. We jointly optimize the Gaussian parameters, the guide/residual codes/decoders, and the appearance decoder with Adam optimizer using a photometric $l_1$ loss, a segmentation loss $\mathcal{L}_{seg}$ to match the rendered hair mask to the ground truth masks, and $\mathcal{L}_{dir}$ loss to match the hair orientation, for 10k iterations.
For hair card generation,  we cluster strands into $N_{c}$ groups (800 for real-world hair, 400 for synthetic hair due to typically containing fewer strands). 
We configure CGHair with $N_{T}=64$ hair card clusters, so as the codebooks for all data. Each codebook contains $K=10$ appearance features, with latent dimensionality $D = 64$.
We uniformly sample $N_S = 25{,}000$ strands (100 points each). The trainable parameters include $99 \cdot N_S$ Gaussian opacity, $N_S$ per-strand appearance logits, $N_T$ shared Gaussian appearance textures, and the shared MLP decoder $\phi_\text{dec}$. We jointly optimize them end-to-end with the 3DGS differentiable renderer for 30k iterations using Adam, initializing strand logits and appearance textures randomly, and assigning a higher learning rate to the logits for faster convergence. For stability, we freeze geometric parameters for the first 7k iterations, then continue optimizing strand geometry with a smaller learning rate.

\subsection{Comparisons with State-of-the-art.} \label{sec:comparisons}

\myparagraph{Strands Generation.}
We compare our recovered strands with GaussianHaircut~\cite{zakharov2024human} and GaussianHair~\cite{luo2024gaussianhair} in ~\cref{fig:comparison}. GaussianHaircut generates hair with low density , e.g., the back of the head (third row) and  generates hair with inaccurate geometry (third row) and intersecting strands (first row). GaussianHair is better in preserving fine wisps but generates overall noisier geometry and still exhibits partial hair loss, e.g., back of the head (third row). In contrast, our method demonstrates superior capability in recovering complete and smooth hair geometry, e.g., the hair line in the long hair case  (third row), thanks to the adopted strong hairstyle prior
Moreover, our pipeline is 4× faster than GaussianHaircut and 3x faster than GaussianHair. Besides, unlike GaussianHair, which requires tedious manual work, our method is fully automatic and thus more practical. Check the supplementary material for more details.

\myparagraph{Compact Gaussian Hair.}
\begin{figure}
\vspace{-5pt} 
	\includegraphics[width=1.0\linewidth]{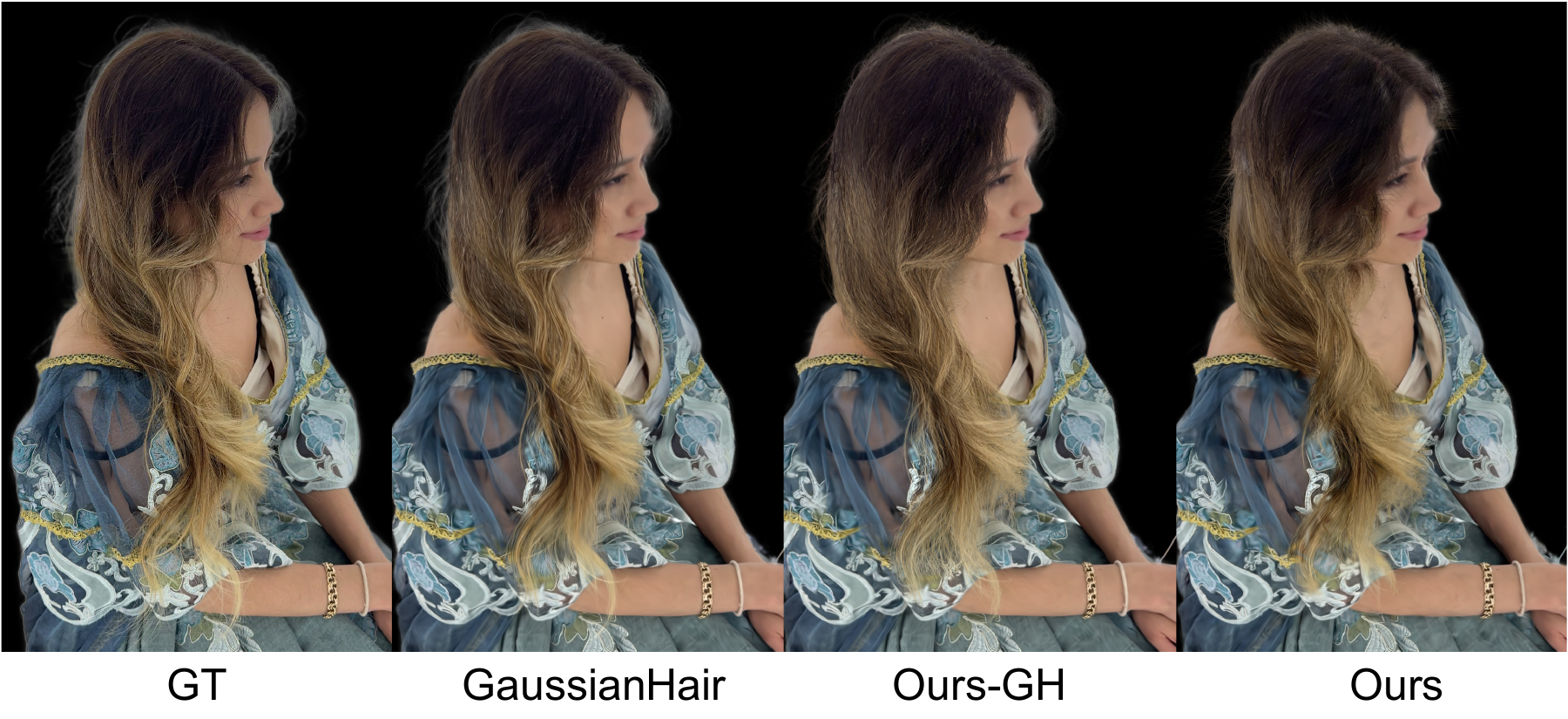}
        \vspace{-20pt} 
	\caption{We integrate our CGHair module with GaussianHair~\cite{luo2024gaussianhair} (Ours-GH), and achieve superior hair details.}
	\label{fig:appearance_comp2}
\vspace{-10pt} 
\end{figure}
\begin{figure*}[t!]
    \vspace{-20pt} 
	\includegraphics[width=\linewidth]{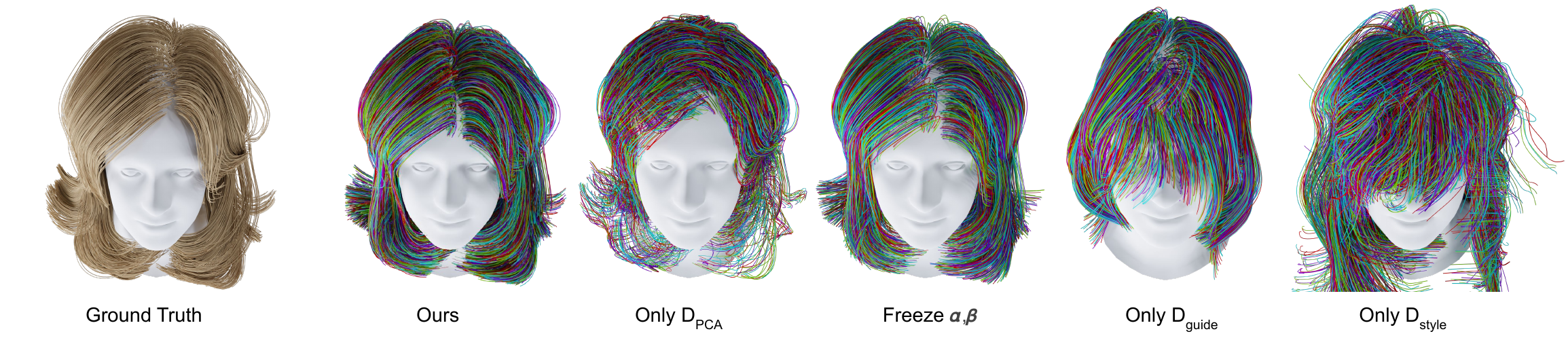}
    \vspace{-20pt} 
	\caption{Strand generator ablations, showing reconstructed hair geometries. Refer to the supplementary material for more results.}
    \vspace{-10pt} 
	\label{fig:synthetic_ablations}
\end{figure*}
We compare CGHair with state-of-the-art hair renderers, GaussianHair~\cite{luo2024gaussianhair} and GaussianHaircut~\cite{zakharov2024human}, using our reconstructed strands for hair-card construction and appearance fitting. 
As shown in ~\cref{fig:appearance_comp}, we showcase several novel views from three challenging cases. GaussianHair offers the most detailed strand-level renderings and preserves flyaway hairs near boundaries, but can over-sharpen (e.g., the side hair in the curly case, first row).
GaussianHaircut produces the smoothest results yet loses strand-level details—especially in the curly (first row) and long straight (third row) cases—and shows forehead artifacts due to strand-geometry misalignment (third row). 
In contrast, our method recovers higher-frequency details in both geometry and appearance, achieving quality comparable to GaussianHair with an appearance model with over 200x compression rate. Check the quantitative results of rendering and compression rate in the supplementary material.
We further incorporate our CGHair as an additional compact module on top of the GaussianHair reconstructed strands, compare the final renderings, as shown in ~\cref{fig:appearance_comp2}. This demonstrates the plug-and-play nature of our method, making it applicable to various types of strands.

\begin{table}
\centering
\vspace{-5pt} 
\caption{Quantitative ablation studies on strand generation.}
\vspace{-10pt}
\label{tab:strand_ablation}
\resizebox{1.0\linewidth}{!}{%
\begin{tabular}{cccccc}
\hline
Metric & Ours & Only $\mathcal{D}_{PCA}$ & Freeze $\alpha,\beta$ & Only $\mathcal{D}_{style}$ & Only $\mathcal{D}_{guide}$  \\
\hline
Pos. Error $\downarrow$ & \textbf{0.148}  &  0.162   & \underline{0.159}  & 0.182   & 0.170 \\
Cur. Error $\downarrow$ & \textbf{6.73}   & 9.12     & \underline{6.98}   & 7.48     & 7.19 \\

\hline
\end{tabular}
}
\vspace{-10pt} 
\end{table}
\subsection{Ablation Studies.} \label{sec:ablations}

We conduct ablation studies to analyze the impact of each module on strand generation and appearance modeling.

\myparagraph{Strand generation.}
We ablate over each component of our strand generator on a synthetic hairstyle from the HairSalon dataset~\cite{hu2015single} by measuring positional error (mean $l_2$ distance) and curvature error ($l_{1}$ norm of curvature) between generated and ground truth strands.
As shown in ~\cref{tab:strand_ablation}, \textit{Only $\mathcal{D}_{PCA}$} denotes an implementation based on GaussianHaircut~\cite{zakharov2024human} where only its MLP strand decoder is replaced with the PCA-based decoder; its worst performance shows the importance of our additional modules. 
\textit{Freeze $\alpha,\beta$} uses our full model but fixes the hair texture codes $\alpha$ and $\beta$ while optimizing the remaining components. \textit{Only $\mathcal{D}_{style}$} and \textit{Only $\mathcal{D}_{guide}$} optimize only the corresponding guide and style texture generator while freezing other components. Compared to our full model, these results demonstrate that each component is crucial for high-quality and accurate strand generation. 
We show qualitative results on a synthetic sample in ~\cref{fig:synthetic_ablations}, and refer to the supplementary material for additional results on real hair samples.

\begin{table}
\small
\centering
\vspace{-5pt}
\caption{Quantitative ablation on key modules of CGHair.}
\vspace{-10pt}
\resizebox{1.0\linewidth}{!}{
\begin{tabular}{lccccc}
\toprule
Method & PSNR~$\uparrow$ & SSIM~$\uparrow$ & LPIPS~$\downarrow$ & Size (MB)~$\downarrow$ & Ratio~$\uparrow$ \\
\midrule
Unique                & \textbf{33.99} & \textbf{0.982} & \textbf{0.019} & 163.7 & 1.0 \\
W/o card clustering   & 28.00 & 0.938 & 0.042 & 22.42 & 7.30 \\
Single-SH             & 30.48 & 0.958 & 0.037 & \textbf{0.46}  & \textbf{355.9} \\
Multi-SH              & \underline{33.50} & \underline{0.978} & \underline{0.021} & 11.95 & 13.70 \\
Latents(full)         & 32.40 & 0.970 & 0.026 & \underline{0.71}  & \underline{230.6} \\
\bottomrule
\end{tabular}
}
\label{tab:component_ablation}
\vspace{-10pt}

\end{table}

\begin{table}
\centering
\tiny
\vspace{-0pt}
\caption{Codebook size $K$ ablations.}
\vspace{-10pt}
\resizebox{\linewidth}{!}{
\begin{tabular}{lccccc}
\toprule
$K$ & PSNR~$\uparrow$ & SSIM~$\uparrow$ & LPIPS~$\downarrow$ & Size (MB)~$\downarrow$ & Ratio~$\uparrow$ \\
\midrule
10 & \textbf{32.40} & \textbf{0.970} & 0.026 & \textbf{0.71} & \textbf{230.6} \\
30 & \textbf{32.40} & \textbf{0.970} & 0.026 & \underline{1.72} & \underline{95.17} \\
60 & \underline{32.36} & \underline{0.969} & 0.026 & 3.22 & 50.84 \\
90 & 32.10 & 0.968 & 0.026 & 4.72 & 34.68 \\
\bottomrule
\end{tabular}
}
\label{tab:codebook_size_ablation}
\vspace{-10pt}

\end{table}

\myparagraph{Compact Gaussian Hair ablations.}
We perform ablations with quantitative metrics (PSNR/SSIM/LPIPS), reporting appearance parameter size and compression ratio (vs. \textit{Unique}). The reference model \textit{Unique} is analogous to GaussianHair, assigning each strand an independent set of 99 SH coefficients. We report appearance size only, since geometry parameters (strand positions and opacities) are fixed across variants at 13.64 MB ($7.7\%$ of the baseline). As summarized in ~\cref{tab:component_ablation}: \textit{W/o card clustering} removes card-based clustering and applies a single global shared texture with equal capacity ($K \times D = 640$) to all strands, yielding the worst accuracy and a larger model, underscoring the benefit of card-based clustering under tight parameter budgets. \textit{Single-SH} replaces latent features with one per-strand SH set and drops the MLP decoder, achieving the highest compression but reduced rendering flexibility. \textit{Multi-SH} maps each codebook entry to multiple per-Gaussian SH sets, attaining \textit{Unique}-level quality but exceeding the full model in size. In contrast, the full model (\textit{Latent}) uses shared latent features with a lightweight decoder, compactly encoding appearance and delivering the best quality–size trade-off. The complete qualitative results can be found in the supplementary material.

\begin{table}
\tiny
\centering
\vspace{-5pt}
\caption{Latent strand feature dimensionality $D$ ablations.}
\vspace{-10pt}
\resizebox{\linewidth}{!}{
\begin{tabular}{lccccc}
\toprule
$D$ & PSNR~$\uparrow$ & SSIM~$\uparrow$ & LPIPS~$\downarrow$ & Size (MB)~$\downarrow$ & Ratio~$\uparrow$ \\
\midrule
32  & \underline{32.39} & \underline{0.969} & \textbf{0.026} & \textbf{0.60} & \textbf{272.8} \\
64  & \textbf{32.40} & \textbf{0.970} & \textbf{0.026} & \underline{0.71} & \underline{230.6} \\
128 & 32.18 & 0.969 & 0.026 & 0.93 & 176.0 \\
256 & 31.94 & 0.967 & \underline{0.027} & 1.37 & 119.5 \\
\bottomrule
\end{tabular}
}
\label{tab:codebook_dim_ablation}
\vspace{-10pt}

\end{table}

\myparagraph{Codebook Configuration.}
We further study how codebook design impacts quality–compactness. Increasing the codebook size from 10 to 90 entries (~\cref{tab:codebook_size_ablation}) yields only slight fidelity gains but sharply increases storage, reducing compression. Varying the latent dimensionality from 32 to 256 (~\cref{tab:codebook_dim_ablation}) shows that low‑D embeddings preserve comparable visual quality with much lower memory, whereas larger D offers diminishing returns. A compact codebook (e.g., 10 entries with 64‑dim features) thus provides a strong quality–efficiency trade‑off.

\myparagraph{Textures Number.}
We assess how the number of card clusters (textures) affects the quality–storage trade-off (~\cref{tab:texture_number_ablation}). Increasing textures from 16 to 96 steadily improves appearance modeling but moderately enlarges storage, reducing compression. Using 64 clusters offers a good trade-off between fidelity and storage, yielding high fidelity with compact size. In the extreme setting—one texture per hair card (400 total)—performance is best and compression remains high, thanks to parameter sharing among strands within each card and along each strand.

\begin{table}
\tiny
\centering
\vspace{-5pt}
\caption{Quantitative ablations on texture clusters number $N_T$.}
\vspace{-10pt}
\resizebox{1.\linewidth}{!}{
\begin{tabular}{lccccc}
\toprule
$N_T$ & PSNR~$\uparrow$ & SSIM~$\uparrow$ & LPIPS~$\downarrow$ & Size (MB)~$\downarrow$ & Ratio~$\uparrow$ \\
\midrule
16  & 31.49 & 0.964 & 0.030 & \textbf{0.60} & \textbf{272.8} \\
32  & 31.84 & 0.967 & 0.027 & \underline{0.64} & \underline{255.8} \\
64  & 32.40 & 0.970 & 0.026 & 0.71 & 230.6 \\
96  & \underline{32.57} & \underline{0.971} & \underline{0.025} & 0.79 & 207.2 \\
\midrule
400  & \textbf{33.04} & \textbf{0.975} & \textbf{0.022} & 1.53 & 108.0 \\
\bottomrule
\end{tabular}
}
\label{tab:texture_number_ablation}
\vspace{-15pt}
\end{table}

\section{Conclusion}
\label{sec:conclusion}

We present CGHair, a compact Gaussian-based hair representation tailored with a comprehensive pipeline to reconstruct hair from multi-view images. Our approach firstly achieves in reconstructing hair strands rapidly with a generative, prior-driven approach. By grouping the strands into representative hair cards, our method further organizes the hair cards by appearance and proposes shared Gaussian texture codebooks to enable unified appearance attributes modeling. This compact representation integrates seamlessly with the 3DGS pipeline and is optimized with a tailored scheme for multi-view reconstruction. Our method achieves substantially reduced reconstruction time while delivering competitive visual quality with 200x lower memory cost.

{
    \small
    \bibliographystyle{ieeenat_fullname}
    \bibliography{main}
}

\maketitlesupplementary
\section{Additional Results}

\begin{figure*}[!ht]
    \centering
    \includegraphics[width=\linewidth]{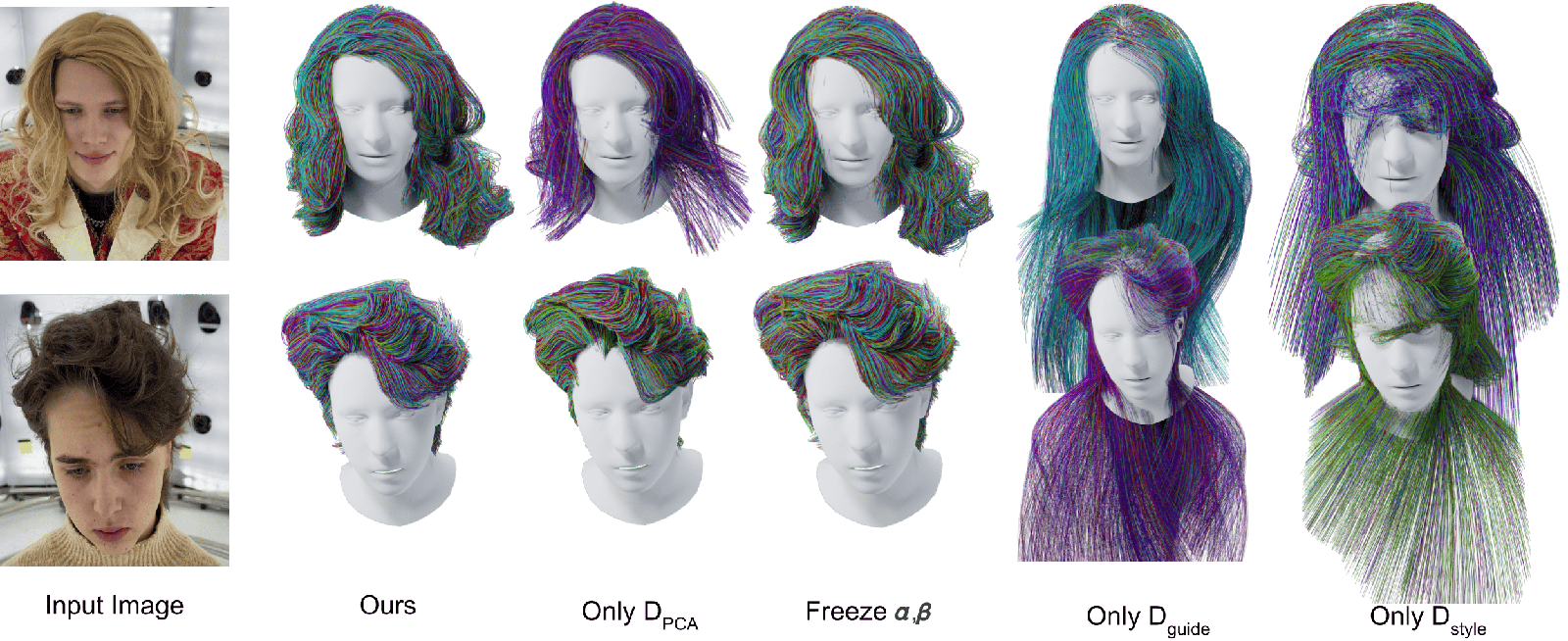}
    \caption{We show the ablations of our strand generator on several real captured hair styles.}
    \label{fig:real_strand_ablations}
\end{figure*}

\subsection{Computational time analysis}

We conduct an ablation study comparing the computational time required by our method with the current state-of-the-art techniques. All experiments are trained on the same setup using a single NVIDIA RTX A4500 GPU.

\paragraph{Strands Reconstruction}

We compare the strand reconstruction time of our approach with the current state-of-the-art methods in~\cref{tab:strand_recon_time}. As shown, our proposed method achieves approximately a fourfold speedup over existing techniques.

\begin{table}[ht]
\centering
\caption{\textbf{Comparison of the strand reconstruction time}. }
\begin{tabular}{c|c|c}
\hline
Model           & Recon. Time  & Ratio   \\ \hline
GaussianHairCut & 11h            & 4.14    \\ 
GaussianHair    & 8h            & 3.01    \\ 
\textbf{Ours}            & \textbf{2h40m}         &\textbf{ 1.00}    \\          
\hline
\end{tabular}
\label{tab:strand_recon_time}
\end{table}

\paragraph{Total end-to-end Reconstruction Time Analysis}
In Tab.~\ref{tab:total_recon_time}, we compare the end-to-end reconstruction time of our method to current state-of-the-art approaches. Our method demonstrates approximately a threefold speedup over previous techniques.

\begin{table}[htbp]
\centering
\caption{\textbf{Comparison on Time for End-to-End Reconstruction}. }
\begin{tabular}{c|c|c}
\hline
Model           & Recon. Time & Ratio   \\ \hline
GaussianHairCut & 15h           & 3.00    \\ 
GaussianHair    & 10.5h          & 2.10    \\ 
\textbf{Ours}            & \textbf{5h}       & \textbf{1.00}   \\          
\hline
\end{tabular}
\label{tab:total_recon_time}
\end{table}

\subsection{Strands reconstruction ablation}
To further evaluate our strand generator, we also conduct the same ablations done in ~\cref{sec:ablations} on real-captured hairstyles. These hairstyles provide a more realistic and challenging test set, enabling us to assess the generalization and effectiveness of our approach more accurately in real-world scenarios as it can bee seen in ~\cref{fig:real_strand_ablations}.

\begin{table}
\centering
\caption{\textbf{Comparison of storage cost}.}
\resizebox{1.0\linewidth}{!}{
\begin{tabular}{c|cc|cc}
\hline
\multicolumn{1}{l|}{} & \multicolumn{2}{c|}{Appearance Features} & \multicolumn{2}{c}{File size}  \\ \hline
Model                 & \multicolumn{1}{c|}{Size (Mb)}  & Ratio  & \multicolumn{1}{c|}{Size (Mb)} & Ratio \\ \hline
GaussianHairCut       & \multicolumn{1}{c|}{544}        & 464    & \multicolumn{1}{c|}{93}        & 77.51   \\
GaussianHair          & \multicolumn{1}{c|}{505}        & 431    & \multicolumn{1}{c|}{90}        & 75   \\
Ours                  & \multicolumn{1}{c|}{1.17}       & 1.00    & \multicolumn{1}{c|}{1.20}      & 1.00   \\ \hline
\end{tabular}
}
\label{tab:storage}
\end{table}

\definecolor{firstred}{rgb}{0.9, 0.2, 0.18}
\definecolor{secondblue}{rgb}{0, 0.6, 0.8}
\definecolor{firstcell}{rgb}{0.94, 0.86, 0.38}
\definecolor{secondcell}{rgb}{0.97, 0.96, 0.52}
\newcommand{\fm}{\cellcolor{firstcell}}
\newcommand{\sm}{\cellcolor{secondcell}}

\begin{table*}[ht!]
\centering
\caption{\textbf{Quantitative comparison on appearance}. Across
three hair types (Curly, Short, and Long) from GaussianHair dataset.}
\resizebox{1\linewidth}{!}{
\begin{tabular}{l|ccc|ccc|ccc}
 & \multicolumn{3}{c|}{Curly} & \multicolumn{3}{c|}{Short} & \multicolumn{3}{c}{Long} \\
 \hline

Method & PSNR$\uparrow$ & SSIM$\uparrow$ & LPIPS$\downarrow$ & PSNR$\uparrow$ & SSIM$\uparrow$ & LPIPS$\downarrow$ & PSNR$\uparrow$ & SSIM$\uparrow$ & LPIPS$\downarrow$ \\ 
\hline
GaussianHairCut         & 30.07 & \sm0.9225 & 0.0565 & 31.03 & \fm0.9041 & 0.0774          & 27.78 & \sm0.8719 & 0.1010\\
GaussianHair            & \fm31.56 & \fm0.9290 & \fm0.0466 & \fm32.07 & \sm0.9020 & \fm0.0597 & \fm29.12 & \fm0.8791 & \fm0.0749   \\ 
\hline
Ours                    & \sm30.34 & 0.9153 & \sm0.0528 & \sm31.15 & \sm0.8928 & \sm0.0696 & \sm28.60 & 0.8696 & \sm0.0815 \\
\end{tabular}
}
\label{tab:comparisons_appearance}
\end{table*}

\begin{figure*}[!ht]
\centering
	\includegraphics[width=1.0\linewidth]{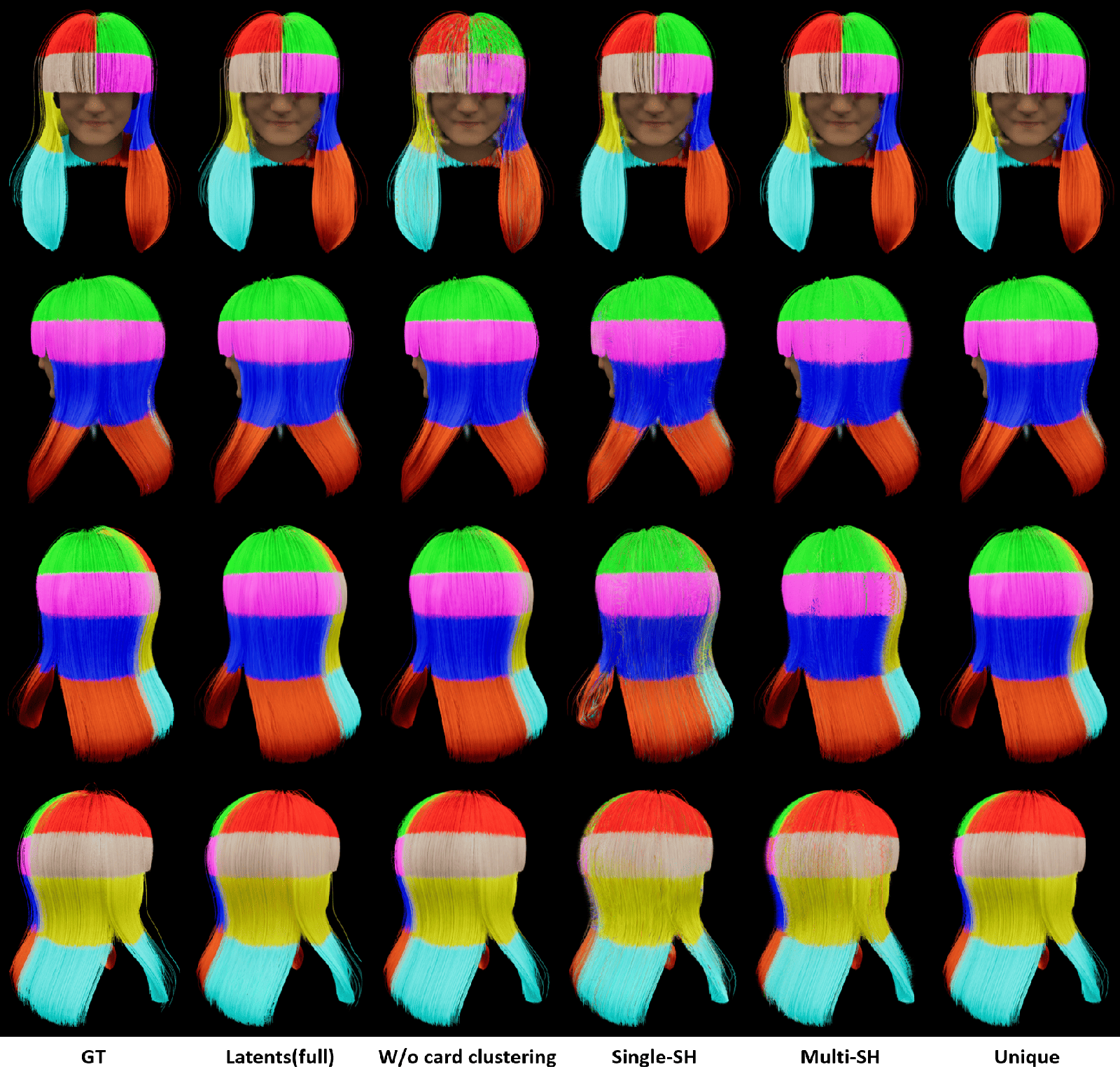}
	\caption{Qualitative ablation results using a carefully designed synthetic hairstyle, illustrating the algorithm's robustness in complex scenarios where hair exhibits multiple color tones.}
	\label{fig:appearance_ablations}
\end{figure*}

\subsection{CGHair}

\paragraph{Memory Footprint Analysis.} 
We present a quantitative analysis of the memory cost of the file size and appearance features of our method in ~\cref{tab:storage}.

\begin{figure*}[t!]
    \includegraphics[width=0.99\linewidth]{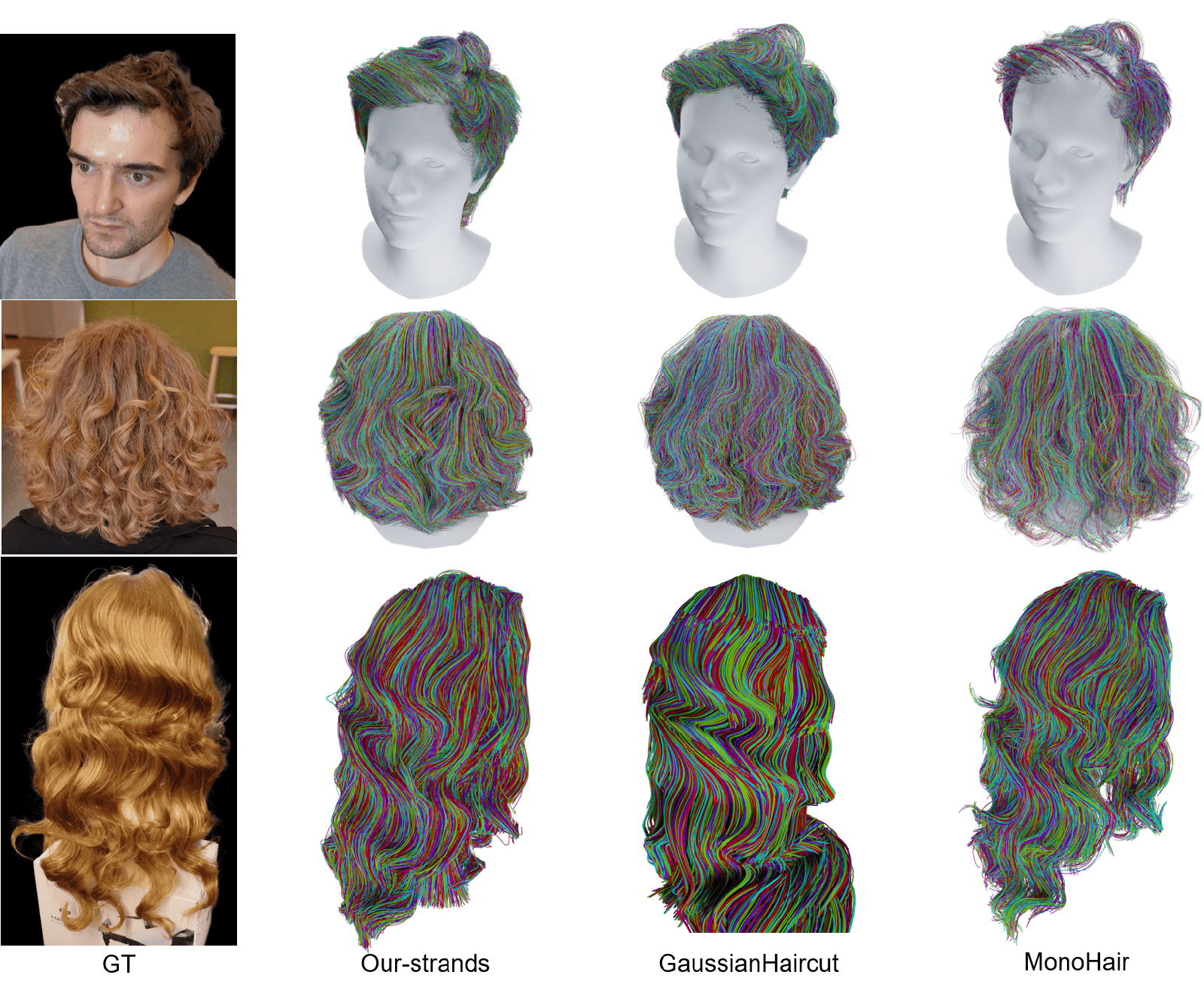} 
    \caption{More hairstyles from the MonoHair dataset. We compare our reconstructed strands against MonoHair and GaussianHaircut.}
    \label{fig:more_strands_results}
\end{figure*}

\begin{figure*}[t!]
    \includegraphics[width=\linewidth]{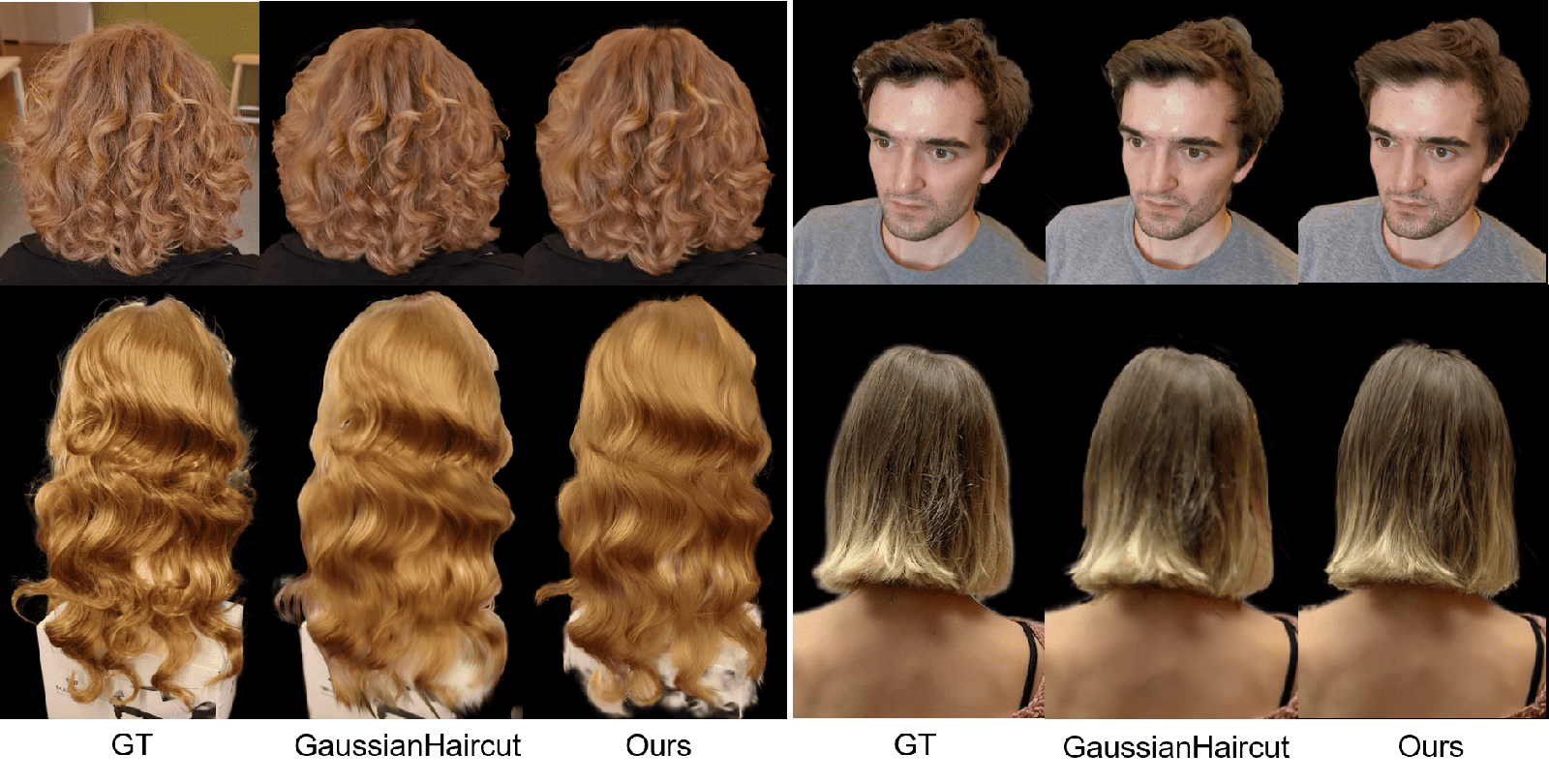} 
    \caption{More Hairstyles appearance rendering comparisons. We compare our appearance rendering results against GaussianHaircut on 3 hairstyles from the MonoHair dataset and one from the H3DS dataset.}
    \label{fig:more_appearance_results}
\end{figure*}

\begin{figure}[t]
\centering
    \includegraphics[width=0.5\textwidth]{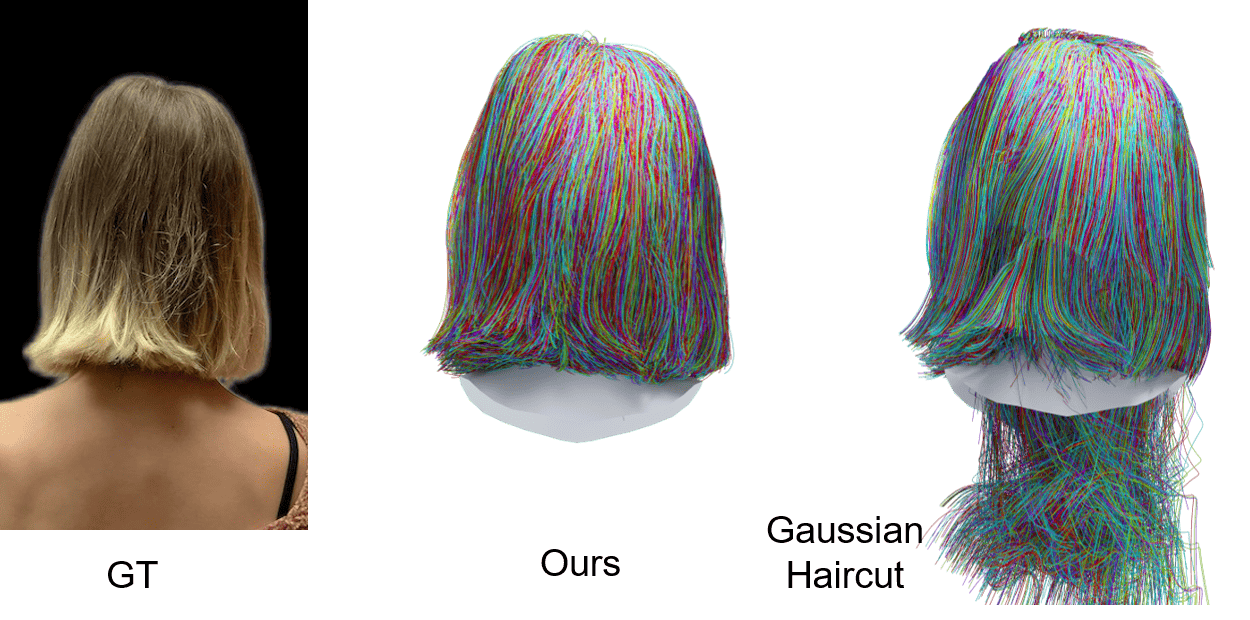} 
    \caption{We also compare our reconstructed strands with GaussianHaircut on the H3DS hairstyle; note that GaussianHaircut failed.}
    \label{fig:h3ds_strands}
\end{figure}

\begin{figure}[t]
\centering
    \includegraphics[width=0.5\textwidth]{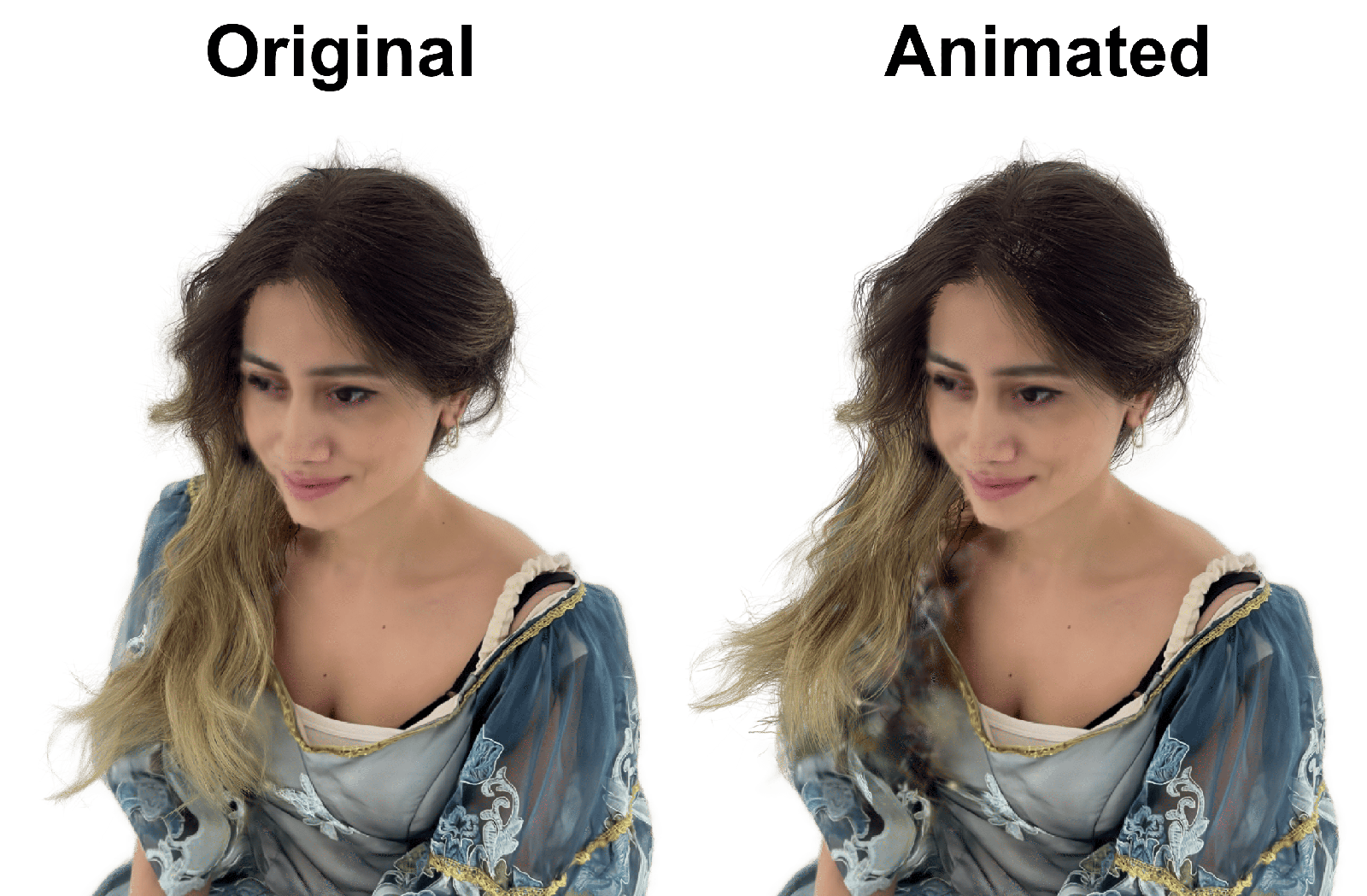} 
    \caption{Animated hair: On the left, original reconstruction, on the right, random timestamp frame from a strand-based animation.}
    \label{fig:animation}
\end{figure}

\paragraph{Quantitative comparisons.}

In ~\cref{tab:comparisons_appearance}, we present a quantitative comparison of the appearance quality of different methods using three common metrics: PSNR (Peak Signal-to-Noise Ratio), SSIM (Structural Similarity Index), and LPIPS (Learned Perceptual Image Patch Similarity).
When comparing the performance of the three methods—GaussianHairCut, GaussianHair, and Ours—across three hair types (Curly, Short, and Long), we observe that GaussianHair performs the best, while our method (Ours) exhibits competitive performance, with slightly lower PSNR and SSIM values than GaussianHair, yet still outperforms GaussianHairCut in most cases. Notably, our method achieves this superior rendering performance with only a fraction of the parameters used by existing state-of-the-art methods. Specifically, we have reduced the parameter count by over 200 times. This demonstrates the efficiency of our approach, achieving high-quality hair strand reconstruction with a drastically smaller model.
\paragraph{Qualitative ablations.}
We show the rendered images using our ablated models in ~\cref{fig:appearance_ablations}. Here, we have carefully designed the rendering scheme for a synthetic hairstyle, where each strand of hair is rendered in multiple segments with different colors, providing an extremely challenging test. This design allows us to sensitively evaluate the model’s ability to preserve fine-grained strand continuity and appearance consistency under complex spatial and color variations.
As shown in ~\cref{fig:appearance_ablations}, we present four unseen views spaced relatively far apart. 

Consistent with the metrics in ~\cref{tab:component_ablation}, \textit{W/o card clustering} exhibits the worst rendering performance. The \textit{Single-SH} model works well on most common monochromatic hair datasets, but it generates many artifacts in this challenging case, for both geometry and appearance. Although \textit{Multi-SH} achieves high rendering metrics, some perspectives exhibit noise, as seen in the last row. As \textit{Unique} is designed analogously to GaussianHair, it demonstrates good rendering performance too; however, the model has a significantly larger number of parameters. In contrast, our method has only 1/200th of the parameters, offering an excellent performance-to-parameter trade-off.

\subsection{Additional Results of Various Hairstyles and free view visualizations}

We provide additional qualitative results for various hairstyles on both the MonoHair~\cite{wu2024monohair} and the H3DS~\cite{ramon2021h3d} datasets. As shown in Fig.~\ref{fig:more_strands_results}, we show our automatically generated results on 3 MonoHair hairstyles. Our results are better than GaussianHaircut, while MonoHair can preserve more details, which is benefited from a manually tuned hair growing algorithm. As shown in Fig.~\ref{fig:more_appearance_results}, we also provide more appearance rendering results of both our method and GaussianHaircut. Our method achieves comparable rendering ability with a much lower memory footprint. In Fig.~\ref{fig:h3ds_strands}, we showcase our generated strands of H3DS data, where GaussianHaircut failed to reconstruct such a hairstyle.

\begin{figure*}[t]
\centering
	\includegraphics[width=\linewidth]{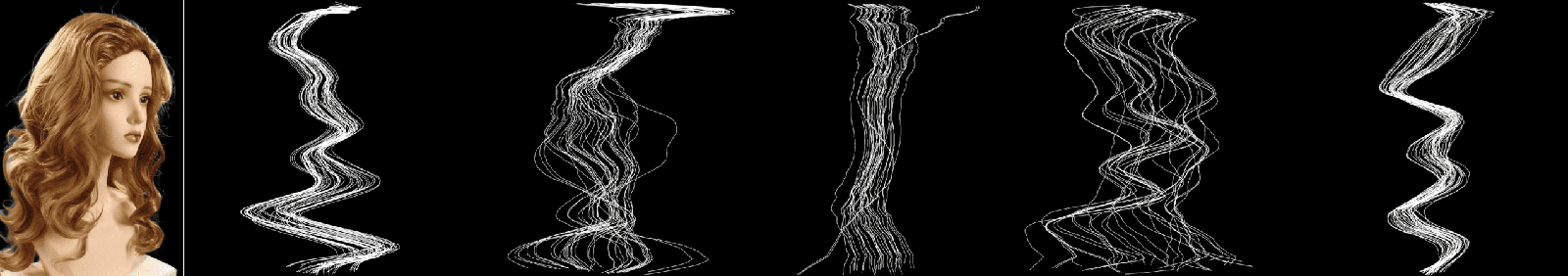}
	\caption{We show several example representative strand textures for grouped strands, which are represented by a hair card. The results are reconstructed from the hairstyle \textit{big wavy} from the Monohair dataset.}
	\label{fig:strands_texture}
\end{figure*}

\paragraph{Visualizations of Hair Card Textures.}
As stated in Sec.~\ref {sec:hair_card}, our method creates hair cards from the grouped strand clusters, and then the hair card textures are generated by 'projecting' the strands onto the hair card mesh faces. We visualize several resulting textures in Fig.~\ref{fig:strands_texture} of a real-captured hairstyle \textit{big wavy}, from the MonoHair~\cite{wu2024monohair} dataset. These textures are then used to cluster the hair cards into similar groups further, to finally generate our shared Gaussian texture codebook.

\subsection{Dynamic Simulation.}

As shown in Fig.~\ref{fig:dynamic}, we show the renderings of the simulated hair strands, for a synthetic hairstyle, the real captured hairstyle \textit{long} from GaussianHair dataset and the hairstyle \textit{big wavy} from MonoHair dataset. Note that we adopt the simulation results from the CG engine and apply them to drive 
our Gaussian hair strands; in this way, we render the strands in the standard Gaussian rendering framework.

\section{Limitation and discussion.}

As a compact Gaussian-based hair representation that can be reconstructed from multi-view static hair images at a superior memory compression ratio, our work still has several limitations:

\paragraph{Failure Cases.} As an image-based hair reconstruction method, our fully automated, efficient hair strands reconstruction method cannot handle extremely challenging hairstyles, like afro-textured or tightly braided hairstyles. As a general limitation, the previous automated 3D Gaussian-based method, e.g., GaussianHaircut~\cite{zakharov2024human}, failed on such challenging cases. While other works, such as GaussianHair~\cite{luo2024gaussianhair} or the NeRF-based method MonoHair~\cite{wu2024monohair}, also failed, even though they require more manual tuning for better results. The main reason is that the complex, intricate hair structure cannot be observed from external views. A possible direction is to collect more complex synthetic hairstyles or real CT hairstyles~\cite{shen2023CT2Hair} to provide better priors for inner hair structure. 

\begin{figure*}[t]
\centering
	\includegraphics[width=\linewidth]{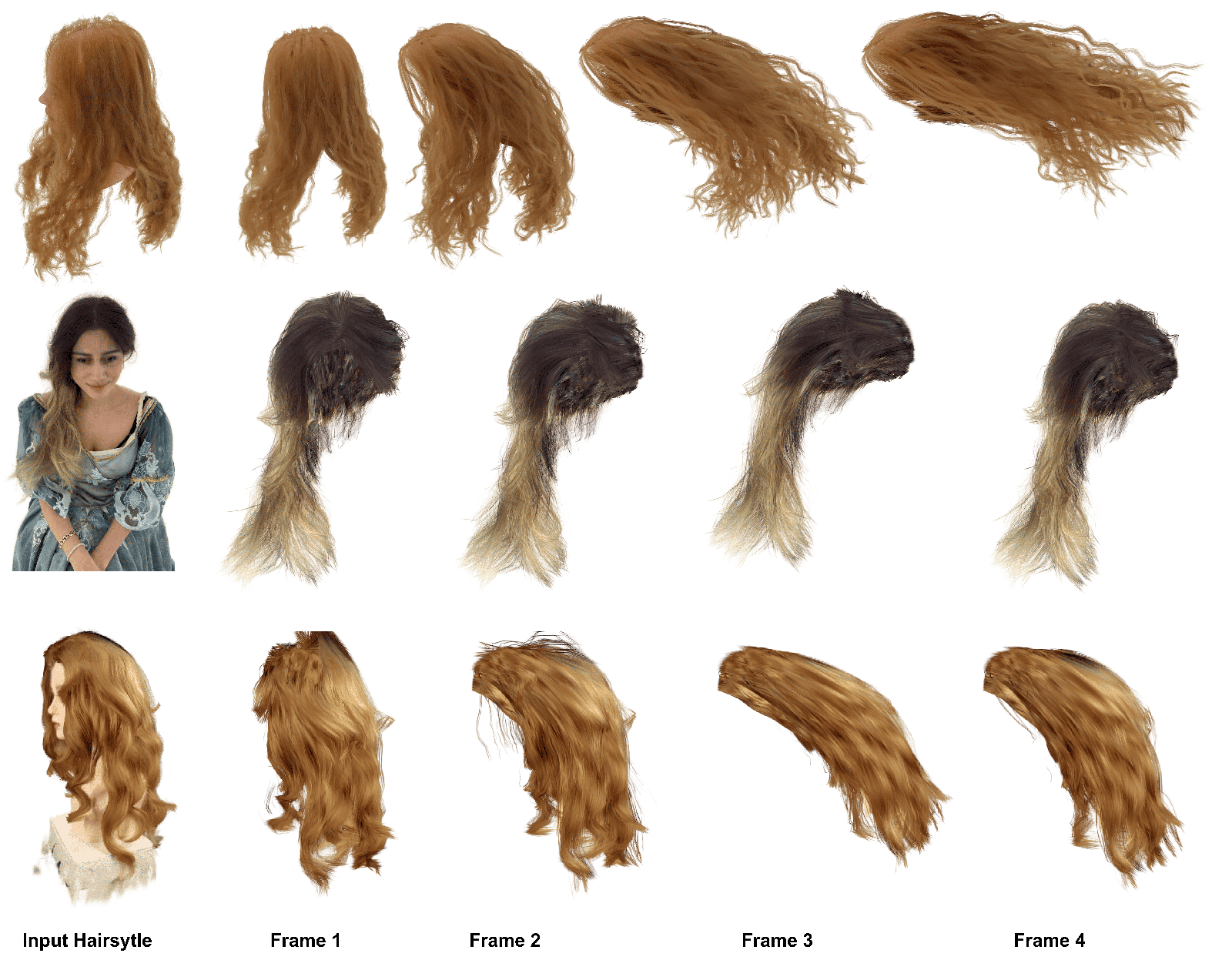}
	\caption{Our explicit strand representation naturally enables strand-based dynamic simulation for a synthetic hairstyle (top), a real captured hairstyle from GaussianHair dataset (middle), and the other one from MonoHair Dataset (bottom). The simulated Gaussian strands are rendered using the standard Gaussian rendering pipeline. Please refer to the project webpage for a better visualization of the dynamic motions.}
	\label{fig:dynamic}
\end{figure*}

\begin{figure*}[t]
\centering
	\includegraphics[width=\linewidth]{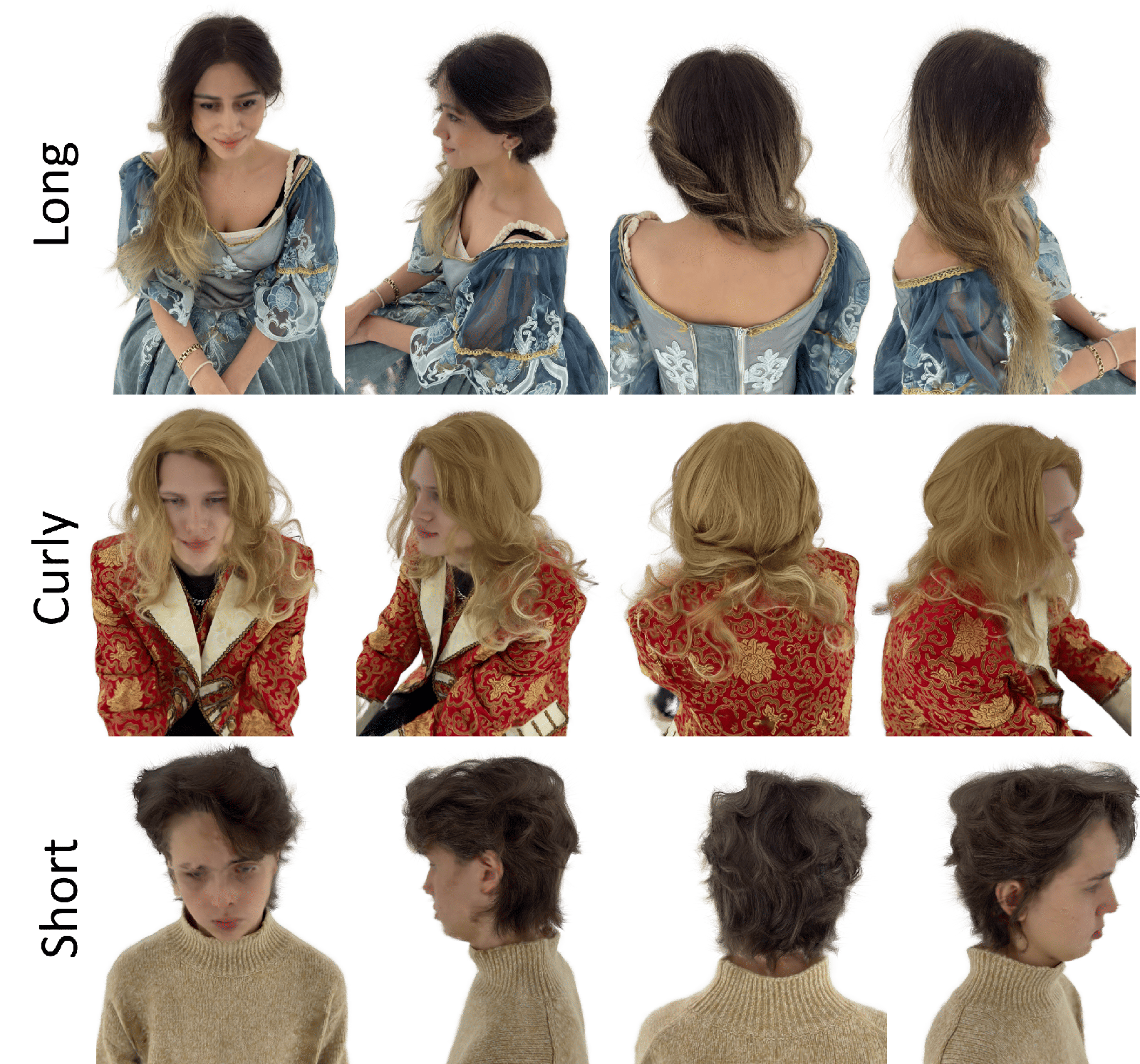}
	\caption{Visualization of our additional free-view renderings of real-captured hairstyles.}
	\label{fig:more_free_views}
\end{figure*}

\paragraph{Dynamic Scenario.}
As our method maintains the explicit hair strands, using hair cards only for clustering and reducing the storage and redundancy of Gaussian-based hair representations, it naturally enables strand-based simulations in CG engines. As shown in Fig.~\ref{fig:animation} and Fig.~\ref{fig:dynamic}, the resulting strand animation from the CG engine can further drive our CGHair. However, the current model is trained only on static monocular captures; the animated results, while keeping visually reasonable hair geometry, may inevitably exhibit artifacts around occluded regions (e.g., occluded body part in Fig.~\ref{fig:animation}) that were never observed during training. Moreover, since Gaussian representations can easily overfit the visible input views, the hair appearance from unseen viewpoints may become less natural. Equipping the current framework with current dynamic techniques and training with dynamically captured multi-view hair video data is a promising direction to mitigate these limitations.

\paragraph{Visual Details Loss.} While our shared Gaussian appearance codebook, driven by card-based hierarchical clustering, achieves over a 200x reduction in the memory footprint of the Gaussian representation, it comes with a trade-off: a slight loss of high-frequency hair details in the appearance rendering. This degradation primarily stems from our current hard-clustering strategy, which remains static once initialized. Therefore, exploring a dynamic or differentiable clustering mechanism that can be jointly optimized during the training process represents a highly desirable direction for future work.

\end{document}